\documentclass[10pt,twocolumn,letterpaper]{article}
\usepackage[pagenumbers]{cvpr}              

\usepackage{graphicx}
\usepackage{amsmath}
\usepackage{amssymb}
\usepackage{booktabs}
\usepackage[table,xcdraw,dvipsnames]{xcolor}
\usepackage{enumitem}
\usepackage{paralist}
\usepackage{bm}
\usepackage{nicefrac}
\usepackage{adjustbox}
\usepackage{bm}
\usepackage[accsupp]{axessibility}

\usepackage[pagebackref=true,breaklinks=true,colorlinks,bookmarks=false,citecolor=blue,linkcolor=blue]{hyperref}

\usepackage[capitalize]{cleveref}
\crefname{section}{Sec.}{Secs.}
\Crefname{section}{Section}{Sections}
\Crefname{table}{Table}{Tables}
\crefname{table}{Tab.}{Tabs.}

\definecolor{iODBlue}{RGB}{220, 232, 250}
\definecolor{gray}{RGB}{238, 238, 238}
\definecolor{columbiablue}{rgb}{0.61, 0.87, 1.0}
\def\newin{\!\in\!}
\def\neweq{\!=\!}
\def\newtimes{\!\times\!}
\definecolor{mygreen}{rgb}{0.4, 0.69, 0.2}
\definecolor{darkyellow}{HTML}{FFA700}
\definecolor{newpurple}{HTML}{BC61F5}

\newcommand*{\affaddr}[1]{\small #1}
\newcommand*{\affmark}[1][*]{\textsuperscript{#1}}

\title{OW-DETR: Open-world Detection Transformer}
\author{
Akshita Gupta\thanks{Equal contribution}~~\affmark[1] \quad
Sanath Narayan\footnotemark[1]~~\affmark[1] \quad
K J Joseph\affmark[2,4] \\
Salman Khan\affmark[4,3]  \quad
Fahad Shahbaz Khan\affmark[4,5] \quad 
Mubarak Shah\affmark[6]\\
\affaddr{\affmark[1]Inception Institute of Artificial Intelligence} \quad
\affaddr{\affmark[2]IIT Hyderabad} \quad
\affaddr{\affmark[3]Australian National University}
\\
\affaddr{\affmark[4]Mohamed Bin Zayed University of Artificial Intelligence} \quad
\affaddr{\affmark[5]CVL, Linköping University} \quad
\affaddr{\affmark[6]University of Central Florida}
}

\begin{document}

\maketitle
\begin{abstract}
    Open-world object detection (OWOD) is a challenging computer vision problem, where the task is to detect a known set of object categories while simultaneously identifying unknown objects. Additionally, the model must incrementally learn new classes that become known in the next training episodes.
    Distinct from standard object detection, the OWOD setting poses significant challenges for generating quality candidate proposals on potentially unknown objects, separating the unknown objects from the background and detecting diverse unknown objects. Here, we introduce a novel end-to-end transformer-based  framework, OW-DETR, for open-world object detection. 
    The proposed OW-DETR comprises three dedicated components namely, attention-driven pseudo-labeling, novelty classification and objectness scoring  to explicitly address the aforementioned OWOD challenges. Our OW-DETR explicitly encodes multi-scale contextual information, possesses less inductive bias, enables knowledge transfer from known classes to the unknown class and can better discriminate between unknown objects and background.
    Comprehensive experiments are performed on two benchmarks: MS-COCO and PASCAL VOC. The extensive ablations reveal the merits of our proposed contributions. Further, %
    our model outperforms the recently introduced OWOD approach, ORE, with absolute gains ranging from  $1.8\%$ to $3.3\%$ in terms of unknown recall on MS-COCO. In the case of incremental object detection, OW-DETR outperforms the state-of-the-art for all settings on PASCAL VOC. Our code is available at \href{https://github.com/akshitac8/OW-DETR}{https://github.com/akshitac8/OW-DETR}.
\end{abstract}

\vspace{-1em}
\section{Introduction\label{sec:intro}}
\vspace{-0.5em}

Open-world object detection (OWOD) relaxes the closed-world assumption in popular benchmarks, where only seen classes appear at inference. Within the OWOD paradigm~\cite{joseph2021towards}, at each training episode, a model learns to detect a given set of \emph{known} objects while simultaneously capable of identifying \emph{unknown} objects. These flagged unknowns can then be forwarded to an oracle (\eg, human annotator), which can label a few classes of interest. Given these \emph{new knowns}, the model would continue updating its knowledge incrementally without retraining from scratch on the previously known classes. This iterative learning process continues in a cycle over the model's life-span.

\begin{figure}
    \centering
    \includegraphics[width=0.95\columnwidth]{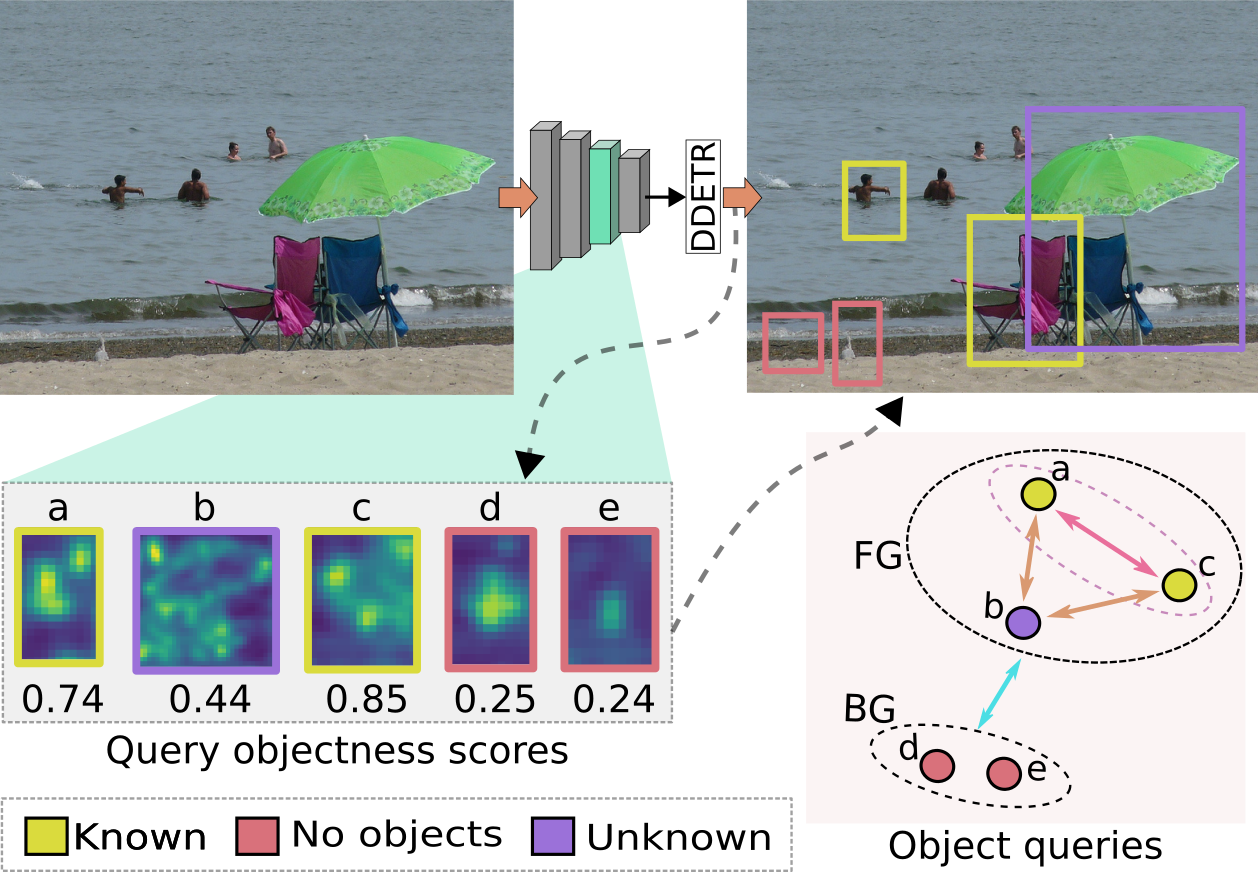}\vspace{-0.2cm}
    \caption{\textbf{Visual illustration of the proposed OW-DETR for open-world object detection (OWOD)}. Here, attention maps obtained from the intermediate features are utilized to score the object queries. The objectness scores of queries are then used to identify the pseudo-unknowns. A separation is enforced between these pseudo-unknowns and ground-truth knowns to detect novel classes. In addition, a separation is also learned between the background and foreground (knowns + unknowns) for effective knowledge transfer from known to unknown class \wrt characteristics of foreground objects. Our OW-DETR explicitly encodes multi-scale context, has less inductive bias, and assumes no supervision for unknown objects, thus well suited for OWOD problem.\vspace{-0.35cm}}
    \label{fig:intro}
\end{figure}

The identification of unknown object classes in OWOD setting poses significant challenges for conventional detectors. \emph{First,} besides an accurate proposal set for seen objects, a detector must also generate quality candidate boxes for potentially unknown objects. \emph{Second,} the model should be able to separate unknown objects from the background utilizing its knowledge about the already seen objects, thereby learning what constitutes a valid object. \emph{Finally}, objects of different sizes must be detected while flexibly modeling their rich context and relations with co-occurring objects.

Recently, the work of~\cite{joseph2021towards} introduces an open-world object detector, ORE, based on the two-stage Faster R-CNN~\cite{ren2015faster} pipeline. Since unknown object annotations are not available during training in the open-world paradigm, ORE proposes to utilize an auto-labeling step to obtain a set of pseudo-unknowns for training. The auto-labeling is performed on class-agnostic proposals output by a region proposal network (RPN). The proposals not overlapping with the ground-truth (GT) known objects but having high `objectness' scores are auto-labeled as unknowns and used in training. These auto-labeled unknowns are then utilized along with GT knowns to perform latent space clustering. Such a clustering attempts to separate the multiple known classes and the unknown class in the latent space and aids in learning a prototype for the unknown class. Furthermore, ORE learns an energy-based binary classifier to distinguish the unknown class from the class-agnostic known class.

While being the first to introduce and explore the challenging OWOD problem formulation, ORE suffers from several issues. (i) ORE relies on a held-out validation set with weak supervision for the unknowns to estimate the distribution of novel category in its energy-based classifier. (ii) To perform contrastive clustering, ORE learns the unknown category with a single latent prototype, which is insufficient to model the diverse intra-class variations commonly present in the unknown objects. Consequently, this can lead to a sub-optimal separation between the knowns and unknowns. (iii) ORE does not explicitly encode long-range dependencies due to a convolution-based design, crucial to capture the contextual information in an image comprising diverse objects.
Here, we set out to alleviate the above issues for the challenging OWOD problem formulation.

\noindent \textbf{Contributions:} Motivated by the aforementioned observations, we introduce a multi-scale context aware detection framework, based on vision transformers \cite{zhu2020deformable}, with dedicated components to address open-world setting including attention-driven pseudo-labeling, novelty classification and objectness scoring for effectively detecting unknown objects in images (see Fig.~\ref{fig:intro}). Specifically, in comparison to the recent OWOD approach ORE %
\cite{joseph2021towards}, that uses a two-stage CNN pipeline, ours is a single-stage framework based on transformers that require less inductive biases and can encode long-term dependencies at multi-scales to enrich contextual information. Different to ORE, which relies on a held-out validation set for estimating the distribution of novel categories, our setting assumes no supervision given for the unknown and is closer to the true open-world scenario. Overall, our novel design offers more flexibility with broad context modeling and less assumptions to address the open-world detection problem. Our  main contributions are:
\begin{compactitem}
    \item We propose a transformer-based open-world detector, OW-DETR, that better models the context with mutli-scale self-attention and deformable receptive fields, in addition to fewer assumptions about the open-world setup along with reduced inductive biases.
    \item We introduce an attention-driven pseudo-labeling scheme for selecting the object query boxes having high attention scores but not matching any known class box as unknown class. The pseudo-unknowns along with the ground-truth knowns are utilized to learn a novelty classifier to distinguish the unknown objects from the known ones. 
    \item We introduce an objectness branch to effectively learn a separation between foreground objects (knowns, pseudo-unknowns) and the background by enabling knowledge transfer from known classes to the unknown class \wrt the characteristics that constitute a foreground object. 
    \item Our extensive experiments on two popular benchmarks demonstrate the effectiveness of the proposed OW-DETR. Specifically, OW-DETR outperforms the recently introduced ORE for both OWOD and incremental object detection tasks. On MS-COCO, OW-DETR achieves absolute gains ranging from $1.8\%$ to $3.3\%$ in terms of unknown recall over ORE.

\end{compactitem}

\begin{figure*}
    \centering
    \includegraphics[width=0.95\textwidth]{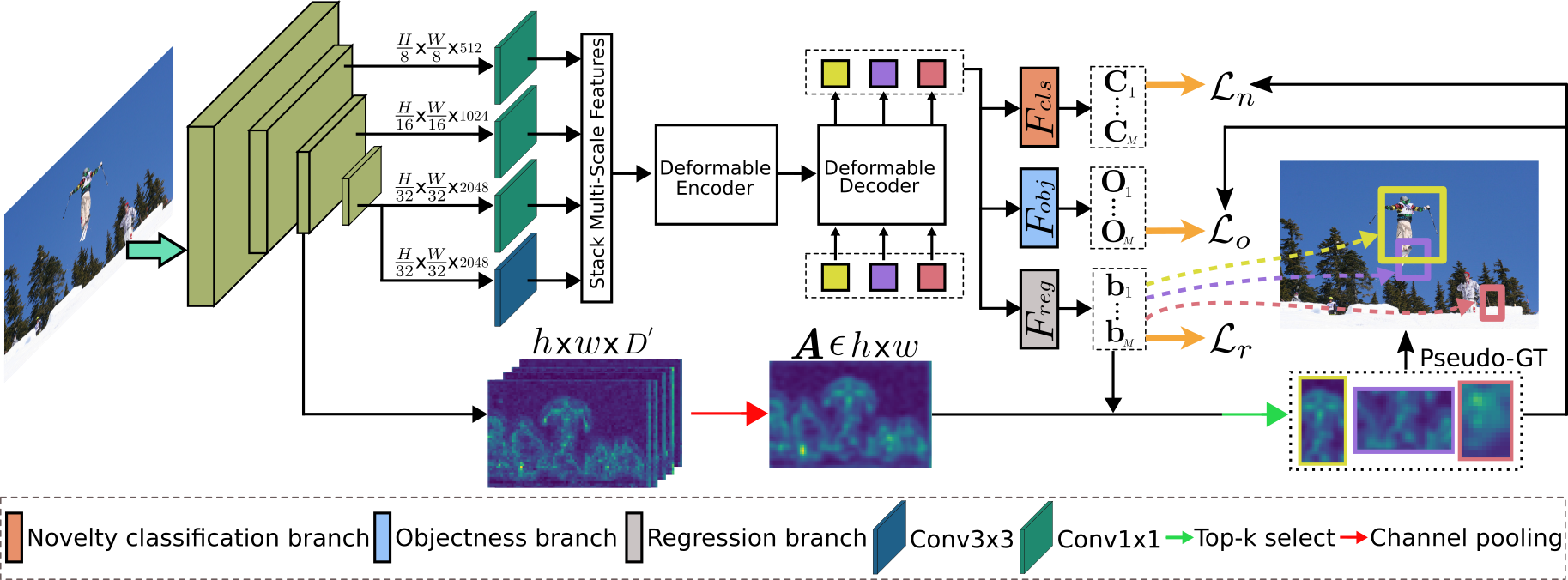}\vspace{-0.25cm}
    \caption{\textbf{Proposed OW-DETR framework.} Our approach adapts the standard Deformable DETR for the OWOD problem formulation by introducing (i) an attention driven pseudo-labeling scheme to select the candidate unknown queries, (ii) a novelty classification branch $F_{cls}$ to distinguish the pseudo unknowns from each of the known classes and (iii) an objectness branch $F_{obj}$ that learns to separate foreground objects (known + pseudo unknowns) from the background. In our OW-DETR, $D$-dimensional multi-scale features for an image $I$ are extracted from the backbone and input to the deformable encoder-decoder along with a set of $M$ learnable object queries $\bm{q} \in \mathbb{R}^D$ to the decoder. At the decoder output, each object query embedding $\bm{q}_e \in \mathbb{R}^D$ is input to three different branches: box regression, novelty classification and objectness. The box co-ordinates are output by the regression branch $F_{reg}$. The objectness branch outputs the confidence of a query being a foreground object, whereas the novelty classification branch classifies the query into one of the known and unknown classes. 
    Our OW-DETR is jointly learned end-to-end with novelty classification loss $\mathcal{L}_n$,  objectness loss $\mathcal{L}_o$ and box regression loss $\mathcal{L}_r$. 
    \vspace{-0.25cm}}
    \label{fig:overall_arch}
\end{figure*}

\section{Open-world Detection Transformer}
\noindent\textbf{Problem Formulation:} Let $\mathcal{K}^t\neweq\{1,2,\cdots,C\}$ denote the set of known object categories at time $t$. Let $\mathcal{D}^t \neweq \{\mathcal{I}^t, \mathcal{Y}^t\}$ be a dataset containing $N$ images $\mathcal{I}^t \neweq \{I_1, \cdots, I_N\}$ with corresponding labels $\mathcal{Y}^t \neweq \{\bm{Y}_1,\cdots,\bm{Y}_N\}$. Here, each $\bm{Y}_i \neweq \{\bm{y}_1,\cdots,\bm{y}_K\}$ denotes the labels of a set of $K$ object instances annotated in the image with $\bm{y}_k \neweq [l_k, x_k, y_k, w_k, h_k]$, where $l_k \newin \mathcal{K}^t$ is the class label for a bounding box represented by $x_k,y_k,w_k,h_k$. Furthermore, let $\mathcal{U} \neweq \{C\!+\!1,\cdots\}$ denote a set of unknown classes that might be encountered at test time. 

As discussed in Sec.~\ref{sec:intro}, in the open-world object detection (OWOD) setting, a model $\mathcal{M}^t$ at time $t$ is trained to identify an unseen class instance as belonging to the unknown class (denoted by label $0$), in addition to detecting the previously encountered known classes $C$. A set of unknown instances $\bm{U}^t \subset \mathcal{U}$ identified by $\mathcal{M}^t$ are then forwarded to an oracle, which labels $n$ novel classes of interest and provides a corresponding set of new training examples.
The learner then incrementally adds this set of new classes to the known classes such that $\mathcal{K}^{t+1} \neweq \mathcal{K}^t + \{C+1,\cdots,C+n\}$. For the previous classes $\mathcal{K}^t$, only few examples can be stored in a bounded memory, mimicking privacy concerns, limited compute and memory resources in real-world settings. Then, $\mathcal{M}^t$ is incrementally trained, without retraining from scratch on the whole dataset, to obtain an updated model $\mathcal{M}^{t+1}$ which can detect all object classes in $\mathcal{K}^{t+1}$. This cycle continues over the life-span of the detector, which updates itself with new knowledge at every episode without forgetting the previously learned classes.

\subsection{Overall Architecture}
Fig.~\ref{fig:overall_arch} shows the overall architecture of the proposed open-world detection transformer, OW-DETR. The proposed OW-DETR adapts the standard Deformable DETR (DDETR)~\cite{zhu2020deformable} for the problem of open-world object detection (OWOD) by introducing (i) an attention-driven pseudo-labeling mechanism (Sec.~\ref{sec:pseudo_labeling}) for selecting likely unknown query candidates; (ii) a novelty classification branch (Sec.~\ref{sec:novelty_classifier}) for learning to classify the object queries into one of the many known classes or the unknown class; and (iii) an `objectness' branch (Sec.~\ref{sec:fg_objectness}) for learning to separate the foreground objects (ground-truth known and pseudo-labeled unknown instances) from the background.
In the proposed OW-DETR, an image $I$ of spatial size $H\times W$ with a set of object instances $\bm{Y}$ is input to a feature extraction backbone. $D$-dimensional multi-scale features are obtained at different resolutions and input to a transformer encoder-decoder containing multi-scale deformable attention modules. The decoder transforms a set of $M$ learnable object queries, aided by interleaved cross-attention and self-attention modules, to a set of $M$ object query embeddings $\bm{q}_e \newin \mathbb{R}^D$ that encode  potential object instances in the image. 

The $\bm{q}_e$ are then input to three branches: bounding box regression, novelty classification and objectness. While the novelty classification ($F_{cls}$) and objectness ($F_{obj}$) branches are single layer feed-forward networks (FFN), the regression branch $F_{reg}$ is a 3-layer FFN. A bipartite matching loss, based on the class and box co-ordinate predictions, is employed to select unique queries that best match the ground-truth (GT) known instances. 
The remaining object queries are then utilized to select the candidate unknown class instances, which are crucial for learning in the OWOD setting. To this end, an attention map $\bm{A}$ obtained from the latent feature maps of the backbone is utilized to compute an objectness score $s_o$ for a query $\bm{q}_e$. The score $s_o$ is based on the activation magnitude inside the query's region-of-interest in $\bm{A}$. The queries with high scores $s_o$ are selected as candidate instances and pseudo-labeled as `unknown'. These pseudo-labeled unknown queries along with the collective GT known queries are employed as foreground objects to train the objectness branch. 
Moreover, while regression branch predicts the bounding box, the novelty classification branch classifies a query into one of the many known classes and an unknown class.
The proposed OW-DETR framework is trained end-to-end using dedicated loss terms for novelty classification ($\mathcal{L}_n$), objectness scoring ($\mathcal{L}_o$), in addition to bounding box regression ($\mathcal{L}_r$) in a joint formulation. Next, we present our OW-DETR approach in detail.

\subsection{Multi-scale Context Encoding\label{sec:def_encdec}}

As discussed earlier in Sec.~\ref{sec:intro}, given the diverse nature of unknown objects that can possibly occur in an image, detecting objects of different sizes while encoding their rich context is one of the major challenges in open-world object detection (OWOD). Encoding such rich context requires capturing long-term dependencies from large receptive fields at multiple scales of the image. Moreover, having lesser inductive biases in the framework that make fewer assumptions about unknown objects, occurring during testing, is likely to be beneficial for improving their detection. 

Motivated by the above observations about OWOD task requirements, we adapt the recently introduced single-stage Deformable DETR~\cite{zhu2020deformable} (DDETR), which is end-to-end trainable and has shown promising performance in standard object detection due to its ability to encode long-term multi-scale context with fewer inductive biases. 
DDETR introduces multi-scale deformable attention modules in the transformer encoder and decoder layers of DETR~\cite{carion2020end} for encoding multi-scale context with better convergence and lower complexity. The multi-scale deformable attention module, based on deformable convolution~\cite{dai2017deformable,zhu2019deformable}, only attends to a small fixed number of key sampling points around a reference point. This sampling is performed across multi-scale feature maps and enables encoding richer context over a larger receptive field. For more details, we refer to~\cite{zhu2020deformable,carion2020end}.
Despite achieving promising performance for the object detection task, the standard DDETR is not suited for detecting unknown class instances in the OWOD setting. To enable detecting novel objects, we introduce an attention-driven pseudo-labeling scheme along with novelty classification and objectness branches, as explained next.

\subsection{Attention-driven Pseudo-labeling\label{sec:pseudo_labeling}}
\begin{figure}
    \centering
    \includegraphics[width=\columnwidth]{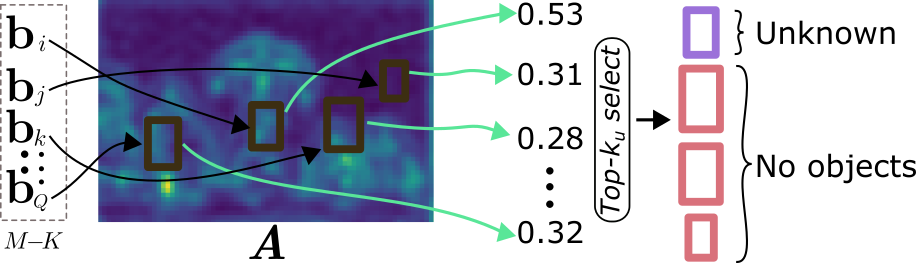}\vspace{-0.2cm}
    \caption{\textbf{An example illustration showing our attention-driven pseudo-labeling}. An objectness score for each of the $M{-}K$ object queries $\bm{q}_e$ is computed as the mean confidence score in a region-of-interest, corresponding to its box proposal $\bm{b}_i$, in the attention feature map $\bm{A}$. A \textit{top}-$k_u$ selection is performed on these $M{-}K$ scores for obtaining $k_u$ pseudo-unknowns.\vspace{-0.3cm}
   }
    \label{fig:pseudo_label}
\end{figure}

For learning to detect unknown objects without any corresponding annotations in the train-set, an OWOD framework must rely on selecting potential unknown instances occurring in the training images and utilizing them as pseudo-unknowns during training. The OWOD approach of ORE~\cite{joseph2021towards} selects proposals having high objectness scores and not overlapping with the ground-truth (GT) known instances as pseudo-unknowns. These proposals obtained from a two-stage detector RPN are likely to be biased to the known classes since it is trained with strong supervision from known classes. Distinct from such a strategy, we introduce a bottom-up attention-driven pseudo-labeling scheme that is better generalizable and applicable in a single-stage object detector. 
Let $\bm{f}$ denote intermediate $D^{'}$-dimensional feature maps extracted from the backbone, with a spatial size $h \newtimes w$. The magnitude of the feature activations gives an indication of presence of an object in that spatial position, and thereby can be used to compute the confidence of objectness within a window. Let $\bm{b}\neweq[x_b,y_b,w_b,h_b]$ denote a box proposal with center $(x_b,y_b)$, width $w_b$ and height $h_b$. The objectness score $s_o(\bm{b})$ is then computed as, 
\begin{equation}
    s_o(\bm{b}) = \frac{1}{h_b\cdot w_b} \sum_{x_b-\frac{w_b}{2}}^{x_b+\frac{w_b}{2}} \sum_{y_b-\frac{h_b}{2}}^{y_b+\frac{h_b}{2}} \bm{A},
\end{equation}
where $\bm{A}\newin\mathbb{R}^{h\times w}$ is the feature map $\bm{f}$ averaged over the channels $D^{'}$. The object proposals in our framework are obtained as the bounding boxes $\bm{b}$ predicted by the regression branch for the $M$ object query embeddings $\bm{q}_e$ output by the deformable transformer decoder. For an image with $K$ known object instances, the objectness score $s_o$ is computed for the $M\!-\!K$ object queries not selected by the bipartite matching loss\footnote{Bipartite matching selects one unique object query per GT instance.} of DDETR as best query matches to the GT known instances. The \textit{top}-$k_u$ queries among $M\!-\!K$ with the high objectness scores $s_o$ are then pseudo-labeled as unknown objects with bounding boxes given by their corresponding regression branch predictions (see Fig.~\ref{fig:pseudo_label}). 

\subsection{Novelty Classification\label{sec:novelty_classifier}}
The ORE~\cite{joseph2021towards} approach introduces an energy-based unknown identifier for classifying a proposal between known and unknown classes. However, it relies on a held-out validation set with weak unknown supervision to learn the energy distributions for the known and unknown classes. In contrast, our OW-DETR does not require any unknown object supervision and relies entirely on the pseudo-unknowns selected using attention-driven pseudo-labeling described in Sec.~\ref{sec:pseudo_labeling}. Furthermore, the classification branch $F_{cls}$ in the standard DDETR classifies an object query embedding $\bm{q}_e$ into one of the known classes or background, \ie, $F_{cls}: \mathbb{R}^D \rightarrow \mathbb{R}^C$. However, when an unknown object is encountered, it fails to classify it into a novel class. To overcome these issues and enable our OW-DETR framework to be trained with only the selected pseudo-unknown objects, we introduce a class label for novel objects in the classification branch. Query embeddings $\bm{q}_e$ selected as pseudo-unknowns are then trained with the pseudo-label (set to $0$ for ease) associated with the novel class in the novelty classification branch $F_{cls}: \mathbb{R}^D \rightarrow \mathbb{R}^{C+1}$. Such an introduction of the novelty class label in classification branch enables $\bm{q}_e$ to be classified as unknown objects in OW-DETR, which otherwise would have been learned as background, as in the standard object detection task. This helps our model to discriminate potential unknown objects from the background.

\subsection{Foreground Objectness\label{sec:fg_objectness}}
As discussed above, the novelty classification branch $F_{cls}$ is class-specific and classifies a query embedding $\bm{q}_e$ into one of the $C+1$ classes: $C$ known classes or $1$ unknown class or background. While this enables the learning of  class-specific separability between known and unknown classes, it does not permit a transfer of knowledge from the known to the unknown objects, which is crucial in understanding as to what constitutes an unknown object in the OWOD setting. Furthermore, the attention-driven pseudo-labeling is likely to be less accurate due to absence of unknown class supervision resulting in most of the query embeddings to be predicted on the background. To alleviate these issues, we introduce a foreground objectness branch $F_{obj}: \mathbb{R}^D \rightarrow [0,1]$ that scores the `objectness'~\cite{redmon2017yolo9000,kong2017ron} of the query embeddings $\bm{q}_e$ in order to better separate the foreground objects (known \textit{and} unknown) from the background. Learning to score the queries corresponding to foreground objects higher than the background enables improved detection of unknown objects which otherwise would have been detected as background. Such a class-agnostic scoring also aids the model to transfer knowledge from the known classes to the unknowns \wrt the characteristics that constitute a foreground object. 

\begin{figure}
    \centering
    \includegraphics[clip=true, trim=1.5em 0em 0em 0em,width=\columnwidth]{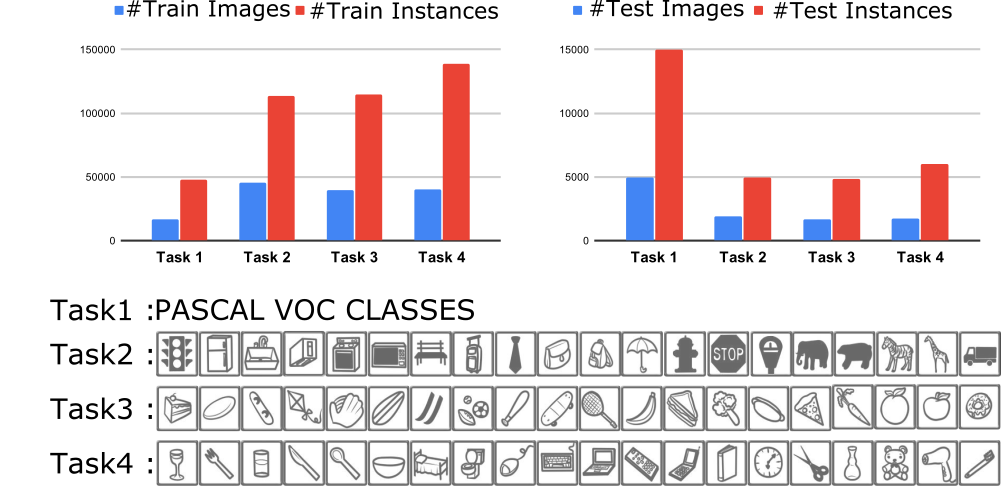}\vspace{-0.3cm}
    \caption{\textbf{Task composition in the OWOD evaluation protocol}. The MS-COCO classes in each task along with the number of images and instances (objects) across splits are shown.\vspace{-0.3cm}}
    \label{fig:data_split}
\end{figure}

\subsection{Training and Inference}
\noindent\textbf{Training:}
Our OW-DETR framework is trained end-to-end using the following joint loss formulation,
\begin{equation}
    \mathcal{L} = \mathcal{L}_n + \mathcal{L}_{r} + \alpha \mathcal{L}_o,
\end{equation}
where $\mathcal{L}_n$, $\mathcal{L}_r$ and $\mathcal{L}_o$ denote the loss terms for novelty classification, bounding box regression and objectness scoring, respectively. While the standard focal loss~\cite{lin2017focal} is employed for formulating $\mathcal{L}_n$  and $\mathcal{L}_o$, the term $\mathcal{L}_r$ is the standard $\ell_1$ regression loss.  
Here, $\alpha$ denotes the weight factor for the objectness scoring.
When a set of new categories are introduced for the incremental learning stage at each episode in OWOD, motivated by the findings in \cite{prabhu2020gdumb,joseph2021towards,wang2020frustratingly}, we employ an exemplar replay based finetuning to alleviate catastrophic forgetting of previously learned  classes. Specifically, the model is finetuned after the incremental step in each episode using a balanced set of exemplars stored for each known class.

\noindent\textbf{Inference:}
$M$ object query embeddings $\bm{q}_e$ are computed for a test image $I$ and their corresponding bounding box and class predictions are obtained, as in~\cite{zhu2020deformable}. Let $C^t$ be the number of known classes at time $t$ in addition to the unknown class, \ie, $C^t \neweq |\mathcal{K}^t|+1$. A \textit{top}-$k$ selection is employed on $M\cdot C^t$ class scores and these selected detections with high scores are used during the OWOD evaluation.

\section{Experiments\label{sec:exp}}

\noindent\textbf{Datasets:} 
We evaluate our OW-DETR on  MS-COCO~\cite{mscoco} for OWOD problem. Classes are grouped into set of non-overlapping tasks $\{T_1,\cdots,T_t,\cdots\}$ {s.t.} classes in a task $T_\lambda$ are not introduced till $t\neweq\lambda$ is reached. While learning for task $T_t$, all the classes encountered in $\{T_\lambda: \lambda\leq t\}$ are considered as \textit{known}. Similarly, classes in $\{T_\lambda: \lambda > t\}$ are considered as \textit{unknown}.
As in~\cite{joseph2021towards}, the $80$ classes of MS-COCO are split into $4$ tasks (see Fig.~\ref{fig:data_split}). The training set for each task is selected from the MS-COCO and Pascal VOC~\cite{everingham2010pascal} train-set images, while Pascal VOC test split and MS-COCO val-set are used for evaluation. \\
\noindent\textbf{Evaluation Metrics:} For known classes, the standard mean average precision (mAP) is used. Furthermore, we use recall as the main metric for unknown object detection instead of the commonly used mAP. This is because all possible unknown object instances in the dataset are not annotated. Recall has been used in~\cite{bansal2018zero,lu2016visual} under similar conditions.\\
\noindent\textbf{Implementation Details:} 
The transformer architecture is similar to DDETR in \cite{zhu2020deformable}.
Multi-scale feature maps are extracted from a ResNet-50~\cite{resnet}, pretrained on ImageNet~\cite{imagenet} in a self-supervised manner~\cite{caron2021emerging}. Such a pretraining mitigates a possible open-world setting violation, which could occur in fully-supervised pretraining (with class labels) due to possible overlap with the novel classes.
The number of queries $M \neweq 100$, while $D \neweq 256$. 
The $k_u$ for selecting pseudo-labels is set to $5$. Moreover, \textit{top}-$50$ high scoring detections per image are used for evaluation during inference. The OW-DETR framework is trained using ADAM optimizer~\cite{kingma2014adam} for $50$ epochs, as in~\cite{zhu2020deformable}. The weight $\alpha$ is set to $0.1$.
Additional details are provided in the appendix.

\begin{table*}[t]
\centering
\caption{\textbf{State-of-the-art comparison for OWOD on MS-COCO.} The comparison is shown in terms of known class mAP and unknown class recall (U-Recall). The unknown recall (U-Recall) metric quantifies a model's ability to retrieve the unknown object instances. 
The standard object detectors (Faster R-CNN and DDETR) in the top part of table achieve promising mAP for known classes but \textit{are inherently not suited for the OWOD setting since they cannot detect any unknown object}.
For a fair comparison in the OWOD setting, we compare with the recently introduced ORE~\cite{joseph2021towards} not employing EBUI. Our OW-DETR achieves improved U-Recall over ORE across tasks, indicating our model's ability to better detect the unknown instances. Furthermore, our OW-DETR also achieves significant gains in mAP for the known classes across the four tasks. Note that since all 80 classes are known in Task 4, U-Recall is not computed. See Sec.~\ref{sec:sota_owod_compare} for more details.  \vspace{-0.3cm}}
\label{tab:sota_owod}
\setlength{\tabcolsep}{3pt}
\adjustbox{width=\textwidth}{
\begin{tabular}{@{}l|cc|cccc|cccc|ccc@{}}
\toprule
 \textbf{Task IDs} ($\rightarrow$)& \multicolumn{2}{c|}{\textbf{Task 1}} & \multicolumn{4}{c|}{\textbf{Task 2}} & \multicolumn{4}{c|}{\textbf{Task 3}} & \multicolumn{3}{c}{\textbf{Task 4}} \\ \midrule
 
 & \cellcolor[HTML]{FFFFED}{U-Recall} & \multicolumn{1}{c|}{\cellcolor[HTML]{EDF6FF}{mAP ($\uparrow$)}} & \cellcolor[HTML]{FFFFED}{U-Recall} & \multicolumn{3}{c|}{\cellcolor[HTML]{EDF6FF}{mAP ($\uparrow$)}} & \cellcolor[HTML]{FFFFED}{U-Recall} & \multicolumn{3}{c|}{\cellcolor[HTML]{EDF6FF}{mAP ($\uparrow$)}} & \multicolumn{3}{c}{\cellcolor[HTML]{EDF6FF}{mAP ($\uparrow$)}}  \\

 & \cellcolor[HTML]{FFFFED}($\uparrow$) & \begin{tabular}[c]{@{}c}Current \\ known\end{tabular} & \cellcolor[HTML]{FFFFED}($\uparrow$) & \begin{tabular}[c]{@{}c@{}}Previously\\  known\end{tabular} & \begin{tabular}[c]{@{}c@{}}Current \\ known\end{tabular} & Both & \cellcolor[HTML]{FFFFED}($\uparrow$) & \begin{tabular}[c]{@{}c@{}}Previously \\ known\end{tabular} & \begin{tabular}[c]{@{}c@{}}Current \\ known\end{tabular} & Both & \begin{tabular}[c]{@{}c@{}}Previously \\ known\end{tabular} & \begin{tabular}[c]{@{}c@{}}Current \\ known\end{tabular} & Both \\ \midrule

Faster-RCNN~\cite{ren2015faster}  & -  & 56.4  & - & 3.7 & 26.7 & 15.2 & - & 2.5 & 15.2 & 6.7 & 0.8 & 14.5 & 4.2 \\ 

\begin{tabular}[c]{@{}l@{}}Faster-RCNN\\ \hspace{1em}+ Finetuning\end{tabular} & \multicolumn{2}{c|}{\begin{tabular}[c]{@{}c}Not applicable in Task 1\end{tabular}}  & - & 51.0 & 25.0 & 38.0  & - & 38.2 & 13.6 & 30.0 & 29.7 & 13.0 & 25.6 \\

DDETR~\cite{zhu2020deformable}  & -  & 60.3 & - & 4.5 & 31.3 & 17.9 & - & 3.3 & 22.5 & 8.5 & 2.5 & 16.4 & 6.0 \\ 

\begin{tabular}[c]{@{}l@{}}DDETR\\ \hspace{1em}+ Finetuning\end{tabular} & \multicolumn{2}{c|}{\begin{tabular}[c]{@{}c}Not applicable in Task 1\end{tabular}}  & - & 54.5 & 34.4 & 44.8  & - & 40.0 & 17.8 & 33.3 & 32.5 & 20.0 & 29.4 \\ 
\midrule
\midrule

ORE $-$ EBUI~\cite{joseph2021towards} & \cellcolor[HTML]{FFFFED} 4.9  & 56.0 & \cellcolor[HTML]{FFFFED}2.9 & 52.7 & 26.0 & 39.4  & \cellcolor[HTML]{FFFFED}3.9 & 38.2 & 12.7 & 29.7 & 29.6 & 12.4 & 25.3 \\ 

\textbf{Ours: OW-DETR} & \cellcolor[HTML]{FFFFED}\textbf{7.5}  & \textbf{59.2} & \cellcolor[HTML]{FFFFED}\textbf{6.2} & \textbf{53.6} & \textbf{33.5} & \textbf{42.9} & \cellcolor[HTML]{FFFFED}\textbf{5.7} & \textbf{38.3} & \textbf{15.8} & \textbf{30.8} & \textbf{31.4} & \textbf{17.1} & \textbf{27.8} \\
\bottomrule

\end{tabular}%
}\vspace{-0.2cm}

\end{table*}

\begin{table*}[t]
\centering
\caption{\textbf{State-of-the-art comparison for incremental object detection (iOD) on PASCAL VOC.} We experiment on 3 different settings. The comparison is shown in terms of per-class AP and overall mAP. The $10$, $5$ and $1$ class(es) in \colorbox{gray}{gray} background are introduced to a detector trained on the remaining $10$, $15$ and $19$ classes, respectively. Our OW-DETR achieves favorable performance in comparison to existing approaches on all the three settings. See Sec.~\ref{sec:ilod_compare} for additional details.\vspace{-0.3cm}}
\label{tab:iOD}
\setlength{\tabcolsep}{3pt}
\adjustbox{width=\textwidth}{%
\begin{tabular}{@{}lccccccccccccccccccccc@{}}
\toprule
{\color[HTML]{009901} \textbf{10 + 10 setting}} & aero & cycle & bird & boat & bottle & bus & car & cat & chair & cow & table & dog & horse & bike & person & plant & sheep & sofa & train & tv & mAP \\ \midrule
ILOD \cite{shmelkov2017incremental} & 69.9 & 70.4 & 69.4 & 54.3 & 48 & 68.7 & 78.9 & 68.4 & 45.5 & 58.1 & \cellcolor[HTML]{EEEEEE}59.7 & \cellcolor[HTML]{EEEEEE}72.7 & \cellcolor[HTML]{EEEEEE}73.5 & \cellcolor[HTML]{EEEEEE}73.2 & \cellcolor[HTML]{EEEEEE}66.3 & \cellcolor[HTML]{EEEEEE}29.5 & \cellcolor[HTML]{EEEEEE}63.4 & \cellcolor[HTML]{EEEEEE}61.6 & \cellcolor[HTML]{EEEEEE}69.3 & \cellcolor[HTML]{EEEEEE}62.2 & 63.2 \\

Faster ILOD \cite{peng2020faster} & 72.8 & 75.7 & 71.2 & 60.5 & 61.7 & 70.4 & 83.3 & 76.6 & 53.1 & 72.3 & \cellcolor[HTML]{EEEEEE}36.7 & \cellcolor[HTML]{EEEEEE}70.9 & \cellcolor[HTML]{EEEEEE}66.8 & \cellcolor[HTML]{EEEEEE}67.6 & \cellcolor[HTML]{EEEEEE}66.1 & \cellcolor[HTML]{EEEEEE}24.7 & \cellcolor[HTML]{EEEEEE}63.1 & \cellcolor[HTML]{EEEEEE}48.1 & \cellcolor[HTML]{EEEEEE}57.1 & \cellcolor[HTML]{EEEEEE}43.6 & 62.1 \\ 
ORE $-$ (CC  + EBUI)~\cite{joseph2021towards} & 53.3 & 69.2 & 62.4 & 51.8 & 52.9 & 73.6 & 83.7 & 71.7 & 42.8 & 66.8 & \cellcolor[HTML]{EEEEEE}46.8 & \cellcolor[HTML]{EEEEEE}59.9 & \cellcolor[HTML]{EEEEEE}65.5 & \cellcolor[HTML]{EEEEEE}66.1 & \cellcolor[HTML]{EEEEEE}68.6 & \cellcolor[HTML]{EEEEEE}29.8 & \cellcolor[HTML]{EEEEEE}55.1 & \cellcolor[HTML]{EEEEEE}51.6 & \cellcolor[HTML]{EEEEEE}65.3 & \cellcolor[HTML]{EEEEEE}51.5 & 59.4 \\
ORE $-$ EBUI~\cite{joseph2021towards} & 63.5 & 70.9 & 58.9 & 42.9 & 34.1 & 76.2 & 80.7 & 76.3 & 34.1 & 66.1 & \cellcolor[HTML]{EEEEEE}56.1 & \cellcolor[HTML]{EEEEEE}70.4 & \cellcolor[HTML]{EEEEEE}80.2 & \cellcolor[HTML]{EEEEEE}72.3 & \cellcolor[HTML]{EEEEEE}81.8 & \cellcolor[HTML]{EEEEEE}42.7 & \cellcolor[HTML]{EEEEEE}71.6 & \cellcolor[HTML]{EEEEEE}68.1 & \cellcolor[HTML]{EEEEEE}77 & \cellcolor[HTML]{EEEEEE}67.7 & 64.5 \\ 
\midrule
\textbf{Ours: OW-DETR} & 61.8 & 69.1 & 67.8 & 45.8 & 47.3 & 78.3 & 78.4 & 78.6 & 36.2 & 71.5 &  \cellcolor[HTML]{EEEEEE} 57.5 &  \cellcolor[HTML]{EEEEEE} 75.3 &  \cellcolor[HTML]{EEEEEE} 76.2 &  \cellcolor[HTML]{EEEEEE} 77.4 &  \cellcolor[HTML]{EEEEEE} 79.5 &  \cellcolor[HTML]{EEEEEE} 40.1 &  \cellcolor[HTML]{EEEEEE} 66.8 &  \cellcolor[HTML]{EEEEEE} 66.3 &  \cellcolor[HTML]{EEEEEE} 75.6 & \cellcolor[HTML]{EEEEEE} 64.1 & \textbf{65.7} \\ \midrule
\midrule

{\color[HTML]{009901} \textbf{15 + 5 setting}} & aero & cycle & bird & boat & bottle & bus & car & cat & chair & cow & table & dog & horse & bike & person & plant & sheep & sofa & train & tv & mAP \\ \midrule
ILOD \cite{shmelkov2017incremental} & 70.5 & 79.2 & 68.8 & 59.1 & 53.2 & 75.4 & 79.4 & 78.8 & 46.6 & 59.4 & 59 & 75.8 & 71.8 & 78.6 & 69.6 & \cellcolor[HTML]{EEEEEE}33.7 & \cellcolor[HTML]{EEEEEE}61.5 & \cellcolor[HTML]{EEEEEE}63.1 & \cellcolor[HTML]{EEEEEE}71.7 & \cellcolor[HTML]{EEEEEE}62.2 & 65.8 \\

Faster ILOD \cite{peng2020faster} & 66.5 & 78.1 & 71.8 & 54.6 & 61.4 & 68.4 & 82.6 & 82.7 & 52.1 & 74.3 & 63.1 & 78.6 & 80.5 & 78.4 & 80.4 & \cellcolor[HTML]{EEEEEE}36.7 & \cellcolor[HTML]{EEEEEE}61.7 & \cellcolor[HTML]{EEEEEE}59.3 & \cellcolor[HTML]{EEEEEE}67.9 & \cellcolor[HTML]{EEEEEE}59.1 & 67.9 \\ 
ORE $-$ (CC  + EBUI)~\cite{joseph2021towards} & 65.1 & 74.6 & 57.9 & 39.5 & 36.7 & 75.1 & 80 & 73.3 & 37.1 & 69.8 & 48.8 & 69 & 77.5 & 72.8 & 76.5 & \cellcolor[HTML]{EEEEEE}34.4 & \cellcolor[HTML]{EEEEEE}62.6 & \cellcolor[HTML]{EEEEEE}56.5 & \cellcolor[HTML]{EEEEEE}80.3 & \cellcolor[HTML]{EEEEEE}65.7 & 62.6 \\
ORE $-$ EBUI~\cite{joseph2021towards} & 75.4 & 81 & 67.1 & 51.9 & 55.7 & 77.2 & 85.6 & 81.7 & 46.1 & 76.2 & 55.4 & 76.7 & 86.2 & 78.5 & 82.1 & \cellcolor[HTML]{EEEEEE}32.8 & \cellcolor[HTML]{EEEEEE}63.6 & \cellcolor[HTML]{EEEEEE}54.7 & \cellcolor[HTML]{EEEEEE}77.7 & \cellcolor[HTML]{EEEEEE}64.6 & 68.5 \\ \midrule
\textbf{Ours: OW-DETR} & 77.1 & 76.5 & 69.2 & 51.3 & 61.3 & 79.8 & 84.2 & 81.0 & 49.7 & 79.6 & 58.1 & 79.0 & 83.1 & 67.8 & 85.4 & \cellcolor[HTML]{EEEEEE}33.2 & \cellcolor[HTML]{EEEEEE}65.1 & \cellcolor[HTML]{EEEEEE}62.0 & \cellcolor[HTML]{EEEEEE}73.9 & \cellcolor[HTML]{EEEEEE}65.0 & \textbf{69.4} \\ \midrule
\midrule

{\color[HTML]{009901} \textbf{19 + 1 setting}} & aero & cycle & bird & boat & bottle & bus & car & cat & chair & cow & table & dog & horse & bike & person & plant & sheep & sofa & train & tv & mAP \\ \midrule
ILOD \cite{shmelkov2017incremental} & 69.4 & 79.3 & 69.5 & 57.4 & 45.4 & 78.4 & 79.1 & 80.5 & 45.7 & 76.3 & 64.8 & 77.2 & 80.8 & 77.5 & 70.1 & 42.3 & 67.5 & 64.4 & 76.7 & \cellcolor[HTML]{EEEEEE}62.7 & 68.2 \\

Faster ILOD \cite{peng2020faster} & 64.2 & 74.7 & 73.2 & 55.5 & 53.7 & 70.8 & 82.9 & 82.6 & 51.6 & 79.7 & 58.7 & 78.8 & 81.8 & 75.3 & 77.4 & 43.1 & 73.8 & 61.7 & 69.8 & \cellcolor[HTML]{EEEEEE}61.1 & 68.5 \\
ORE $-$ (CC  + EBUI)~\cite{joseph2021towards} & 60.7 & 78.6 & 61.8 & 45 & 43.2 & 75.1 & 82.5 & 75.5 & 42.4 & 75.1 & 56.7 & 72.9 & 80.8 & 75.4 & 77.7 & 37.8 & 72.3 & 64.5 & 70.7 & \cellcolor[HTML]{EEEEEE}49.9 & 64.9 \\
ORE $-$ EBUI~\cite{joseph2021towards} & 67.3 & 76.8 & 60 & 48.4 & 58.8 & 81.1 & 86.5 & 75.8 & 41.5 & 79.6 & 54.6 & 72.8 & 85.9 & 81.7 & 82.4 & 44.8 & 75.8 & 68.2 & 75.7 & \cellcolor[HTML]{EEEEEE}60.1 & 68.8 \\ \midrule
\textbf{Ours: OW-DETR} & 70.5 & 77.2 & 73.8 & 54.0 & 55.6 & 79.0 & 80.8 & 80.6 & 43.2 & 80.4 & 53.5 & 77.5 & 89.5 & 82.0 & 74.7 & 43.3 & 71.9 & 66.6 & 79.4 & \cellcolor[HTML]{EEEEEE}62.0 & \textbf{70.2} \\
\bottomrule
\end{tabular}%
}\vspace{-0.5em}

\end{table*}

\begin{figure*}[t]
  \centering\setlength{\tabcolsep}{3pt}
  \resizebox{\textwidth}{!}{%
    \begin{tabular}{cccccc}
      \includegraphics[width=0.3\textwidth, height=0.21\textwidth]{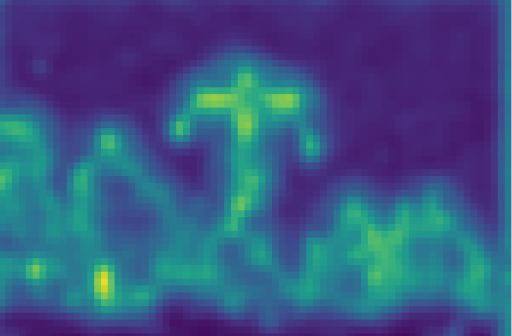}&
      \includegraphics[width=0.3\textwidth, height=0.21\textwidth]{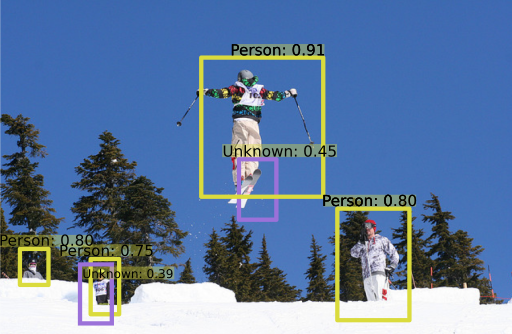}&
      \includegraphics[width=0.3\textwidth, height=0.21\textwidth]{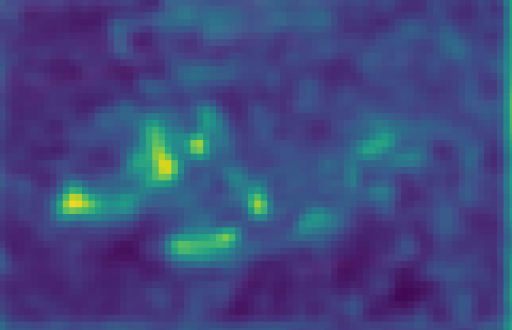}&
      \includegraphics[width=0.3\textwidth, height=0.21\textwidth]{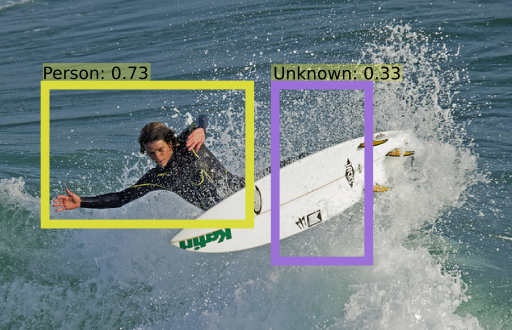}&
      \includegraphics[width=0.3\textwidth, height=0.21\textwidth]{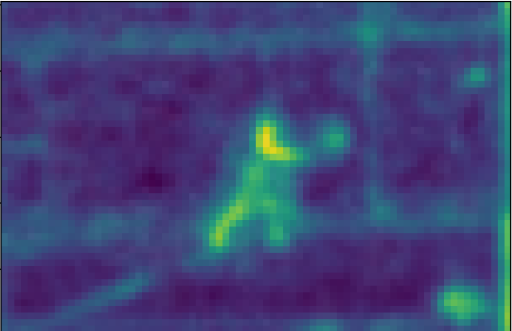}&
      \includegraphics[width=0.3\textwidth, height=0.21\textwidth]{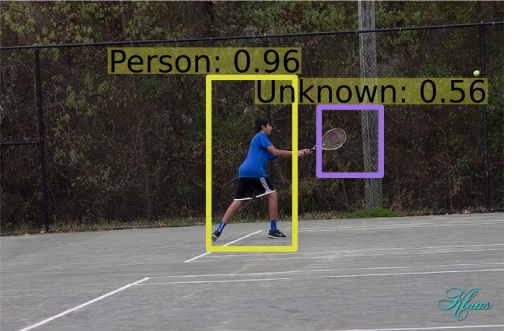}\\

      \includegraphics[width=0.3\textwidth, height=0.21\textwidth]{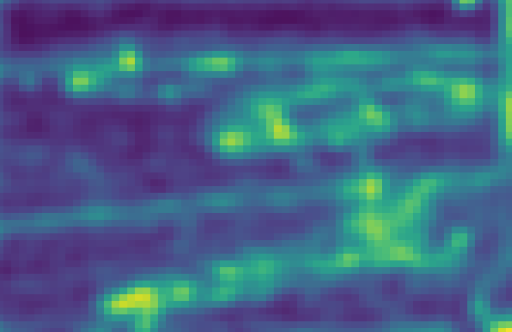}&
      \includegraphics[width=0.3\textwidth, height=0.21\textwidth]{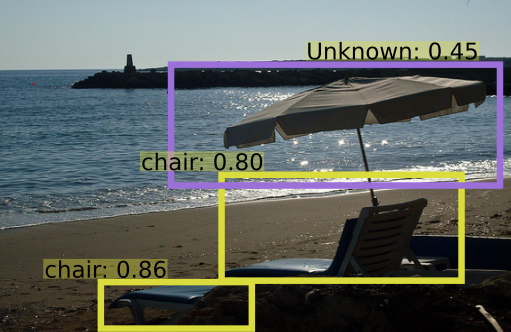}&
      \includegraphics[width=0.3\textwidth, height=0.21\textwidth]{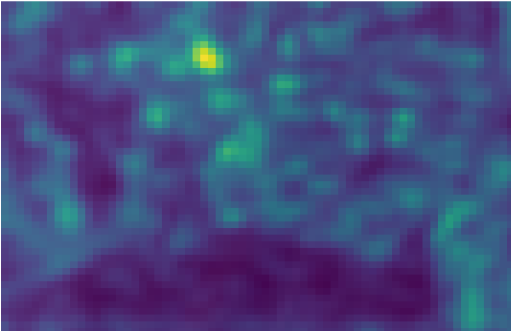}&
      \includegraphics[width=0.3\textwidth, height=0.21\textwidth]{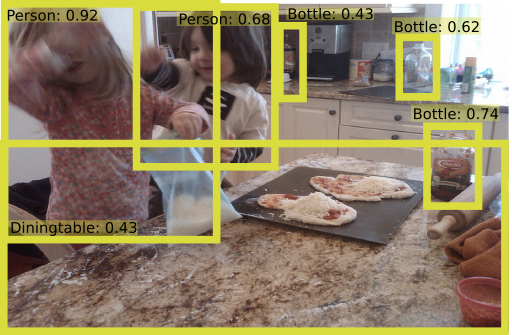}&
      \includegraphics[width=0.3\textwidth, height=0.21\textwidth]{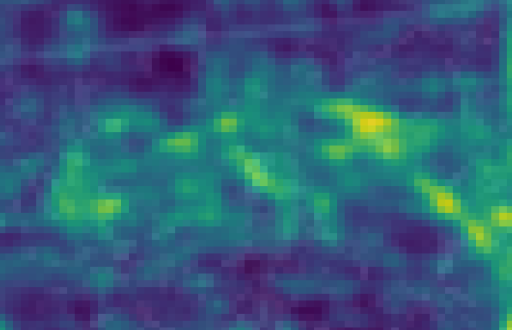}&
      \includegraphics[width=0.3\textwidth, height=0.21\textwidth]{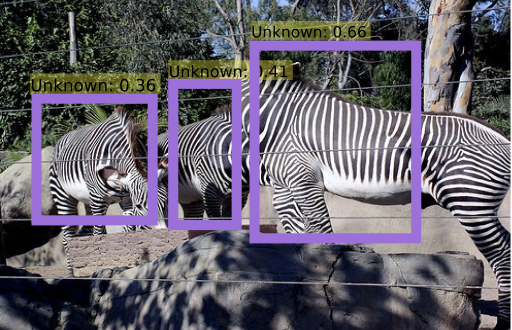}\\
      
      \includegraphics[width=0.3\textwidth, height=0.21\textwidth]{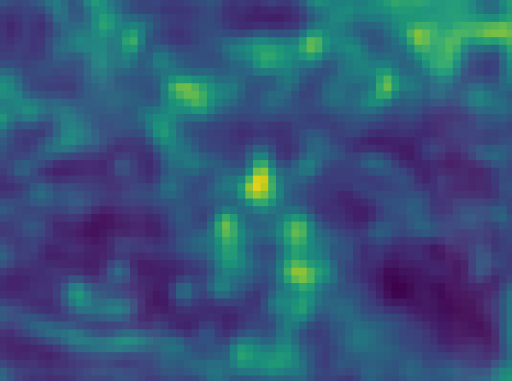}&
      \includegraphics[width=0.3\textwidth, height=0.21\textwidth]{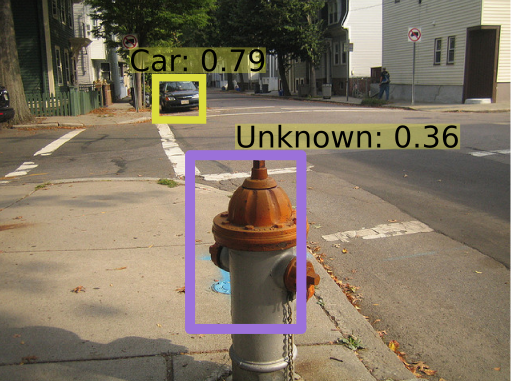}&
      \includegraphics[width=0.3\textwidth, height=0.21\textwidth]{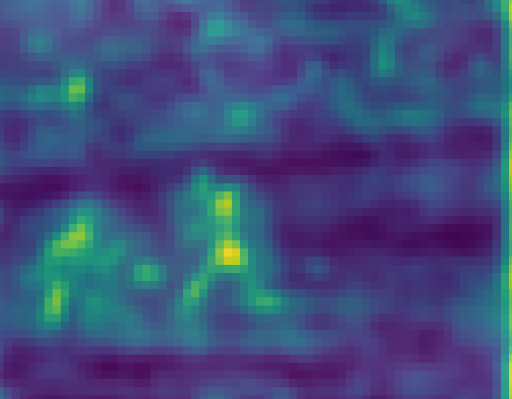}&
      \includegraphics[width=0.3\textwidth, height=0.21\textwidth]{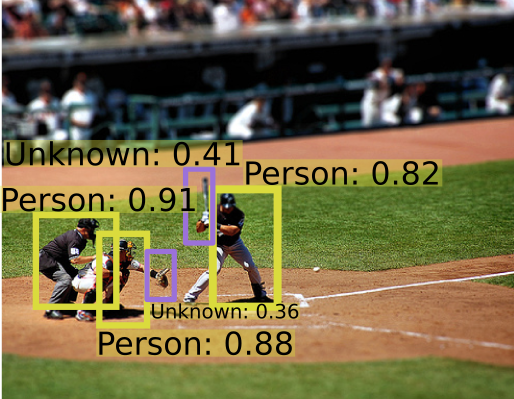}&
      \includegraphics[width=0.3\textwidth, height=0.21\textwidth]{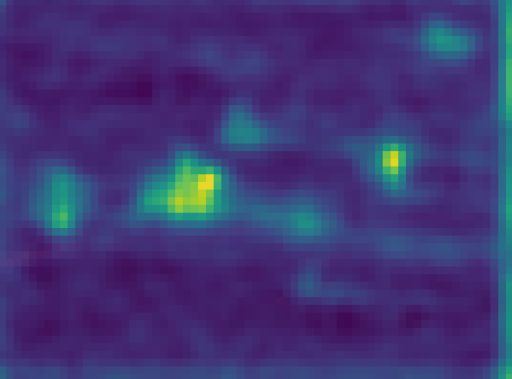}&
      \includegraphics[width=0.3\textwidth, height=0.21\textwidth]{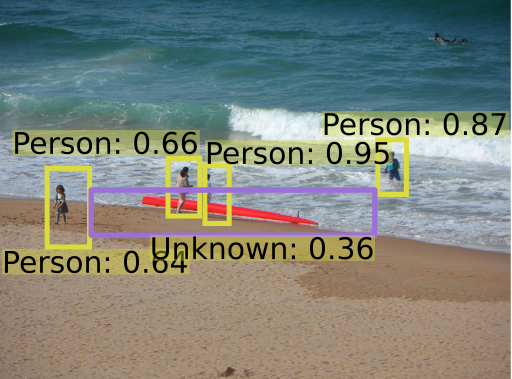}
      \\
      
      \includegraphics[width=0.3\textwidth, height=0.21\textwidth]{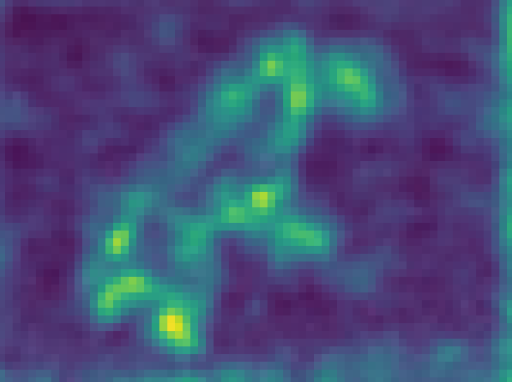}&
      \includegraphics[width=0.3\textwidth, height=0.21\textwidth]{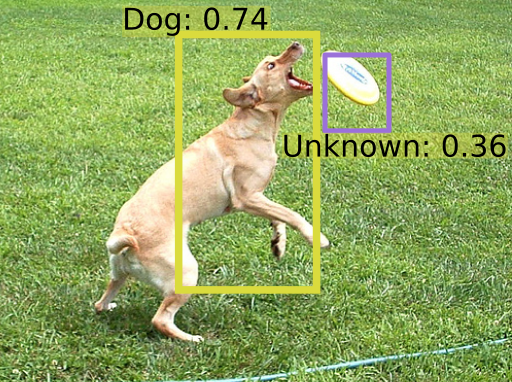}&
      \includegraphics[width=0.3\textwidth, height=0.21\textwidth]{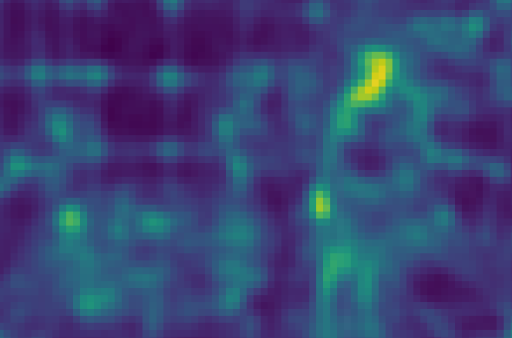}&
      \includegraphics[width=0.3\textwidth, height=0.21\textwidth]{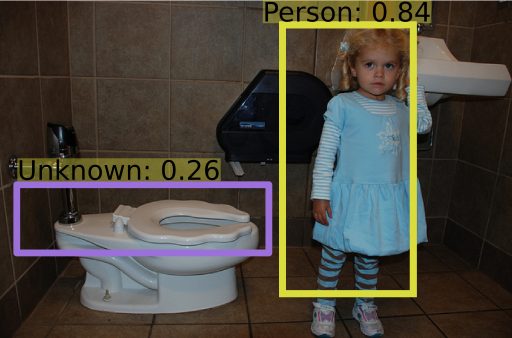}&
      \includegraphics[width=0.3\textwidth, height=0.21\textwidth]{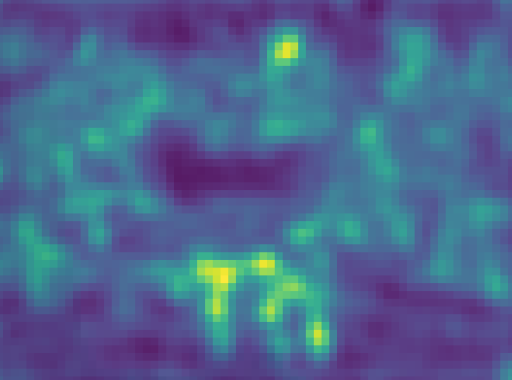}&
      \includegraphics[width=0.3\textwidth, height=0.21\textwidth]{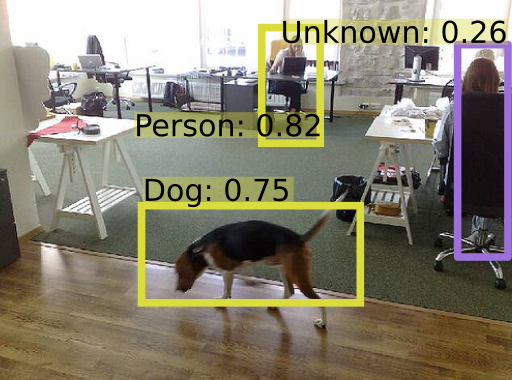}
    
    \end{tabular}}

\vspace{-0.2cm}
 \caption{\textbf{Qualitative results on example images from MS-COCO test set.} For each example image, its corresponding attention map $\bm A$ computed from the intermediate feature maps is shown on its left. The detections obtained from our OW-DETR are overlayed on the known (\textcolor{darkyellow}{yellow}) and unknown (\textcolor{newpurple}{purple}) class objects. We observe that the attention map activations tend to be higher for regions with foreground objects, illustrating the benefits of attention-driven pseudo-labeling for the unknown objects. The unknown objects like \textit{racket} (row 1, right), \textit{umbrella} (row 2, left), \textit{fire hydrant} (row 3, left) are detected reasonably well. Due to the challenging open-world setting, a few unknown objects are missed, \eg, \textit{sink} (row 2, middle), \textit{table} (row 3, right). Nevertheless, these results indicate the promising performance achieved by our OW-DETR framework in the challenging OWOD setting.\vspace{-0.2cm}} 
\label{fig:qual}
\end{figure*}

\subsection{State-of-the-art Comparison\label{sec:sota_owod_compare}}
Tab.~\ref{tab:sota_owod} shows a comparison of our OW-DETR with the recently introduced ORE~\cite{joseph2021towards} on MS-COCO for the OWOD problem. 
We also report the performance of Faster R-CNN~\cite{ren2015faster} and the standard Deformable DETR (DDETR)~\cite{zhu2020deformable} frameworks. The comparison is shown in terms of the known class mAP and unknown class recall (U-Recall). U-Recall quantifies a model's ability to retrieve unknown object instances in the OWOD setting. Note that all 80 classes are known in Task 4 and thereby U-Recall cannot be computed due to the absence of unknown test annotations.
Since both Faster R-CNN and DDETR can only classify objects into known classes but not the unknown, they are \textit{not suited for OWOD setting} and U-Recall cannot be computed for them. 
For a fair comparison in the OWOD setting, we report ORE without its energy-based unknown identifier (EBUI) that relies on held-out validation data with weak unknown object supervision. The resulting ORE$-$EBUI framework achieves U-Recall of $4.9$, $2.9$ and $3.9$ on Task 1, 2 and 3, respectively. Our OW-DETR improves the retrieval of unknown objects, leading to improved performance with significant gains for U-Recall, achieving $7.5$, $6.2$ and $5.7$ on the same tasks 1, 2 and 3, respectively. 
Furthermore, OW-DETR outperforms the best existing OWOD approach of ORE in terms of the known class mAP on all the four tasks, achieving significant absolute gains up to $3.6\%$. While we use the same split as~\cite{joseph2021towards} here for fairness, our OW-DETR also achieves identical gains on a stricter data split (included in appendix) obtained by removing any possible information leakage. The consistent improvement of OW-DETR over ORE, vanilla Faster R-CNN and DDETR emphasizes the importance of proposed contributions towards a more accurate OWOD.

\subsection{Incremental Object Detection\label{sec:ilod_compare}}
As an intuitive consequence of detecting unknown instances, our OW-DETR performs favorably on the incremental object detection (iOD) task. This is due to the decrease in confusion of an unknown object being classified as known class, which enables the detector to incrementally learn the various newer class instances as true foreground objects. Tab.~\ref{tab:iOD} shows a comparison of OW-DETR with existing approaches on PASCAL VOC 2007. As in~\cite{shmelkov2017incremental,peng2020faster}, evaluation is performed on three standard settings, where a group of classes ($10$, $5$ and last class) are introduced incrementally to a detector trained on the remaining classes ($10$, $15$ and $19$). Our OW-DETR performs favorably against existing approaches on all three settings, illustrating the benefits of modeling the unknown object class.

\begin{table*}[t]

\centering
\caption{\textbf{Impact of progressively integrating our contributions into the baseline.} The comparison is shown in terms of known class average precision (mAP) and unknown class recall (U-Recall) on MS-COCO for OWOD setting. Apart from the standard baseline (denoted with $\dagger$), all other models shown include a finetuning step to mitigate catastrophic forgetting. We also show the performance of the oracle (baseline trained with ground-truth unknown class annotations). 
\textit{Although} \texttt{Baseline} \textit{achieves higher mAP for known classes, it is inherently not suited for the OWOD setting since it cannot detect any unknown object}. Integrating the proposed pseudo-labeling based novelty classification (\texttt{NC}) with \texttt{Baseline} enables unknown class detection. Additionally integrating our objectness branch into the framework further improves the retrieval of unknown objects. 
Note that since all 80 classes are known in Task 4, U-Recall is not computed.\vspace{-0.3cm}}
\label{tab:ablation_table}
\setlength{\tabcolsep}{2pt}
\adjustbox{width=\textwidth}{
\begin{tabular}{@{}l|cc|cccc|cccc|ccc@{}}
\toprule

 \textbf{Task IDs} ($\rightarrow$)& \multicolumn{2}{c|}{\textbf{Task 1}} & \multicolumn{4}{c|}{\textbf{Task 2}} & \multicolumn{4}{c|}{\textbf{Task 3}} & \multicolumn{3}{c}{\textbf{Task 4}} \\ \midrule
 
 & \cellcolor[HTML]{FFFFED}{U-Recall} & \multicolumn{1}{c|}{\cellcolor[HTML]{EDF6FF}{mAP ($\uparrow$)}} & \cellcolor[HTML]{FFFFED}{U-Recall} & \multicolumn{3}{c|}{\cellcolor[HTML]{EDF6FF}{mAP ($\uparrow$)}} & \cellcolor[HTML]{FFFFED}{U-Recall} & \multicolumn{3}{c|}{\cellcolor[HTML]{EDF6FF}{mAP ($\uparrow$)}} & \multicolumn{3}{c}{\cellcolor[HTML]{EDF6FF}{mAP ($\uparrow$)}}  \\

 & \cellcolor[HTML]{FFFFED}($\uparrow$) & \begin{tabular}[c]{@{}c}Current \\ known\end{tabular} & \cellcolor[HTML]{FFFFED}($\uparrow$) & \begin{tabular}[c]{@{}c@{}}Previously\\  known\end{tabular} & \begin{tabular}[c]{@{}c@{}}Current \\ known\end{tabular} & Both & \cellcolor[HTML]{FFFFED}($\uparrow$) & \begin{tabular}[c]{@{}c@{}}Previously \\ known\end{tabular} & \begin{tabular}[c]{@{}c@{}}Current \\ known\end{tabular} & Both & \begin{tabular}[c]{@{}c@{}}Previously \\ known\end{tabular} & \begin{tabular}[c]{@{}c@{}}Current \\ known\end{tabular} & Both \\ \midrule

Oracle & \cellcolor[HTML]{FFFFED}31.6  & 62.5  & \cellcolor[HTML]{FFFFED}40.5 & 55.8 & 38.1 & 46.9 & \cellcolor[HTML]{FFFFED}42.6 & 42.4 & 29.3 & 33.9 & 35.6 & 23.1 & 32.5 \\ 
\midrule

\texttt{Baseline}${}^\dagger$  & -  & 60.3 & - & 4.5 & 31.3 & 17.8 & - & 3.3 & 22.5 & 8.5 & 2.5 & 16.4 & 6.0 \\ 

\begin{tabular}[c]{@{}l@{}} \texttt{Baseline} \end{tabular} & \multicolumn{2}{c|}{\begin{tabular}[c]{@{}c}Not applicable in Task 1\end{tabular}}  & - & 54.5 & 34.4 & 44.7  & - & 40.0 & 17.7 & 33.3 & 32.5 & 20.0 & 29.4 \\ 
\midrule

\texttt{Baseline} + \texttt{NC}  & \cellcolor[HTML]{FFFFED}5.9 & 58.1 & \cellcolor[HTML]{FFFFED}4.6 & 52.5 & 32.7 & 42.6  & \cellcolor[HTML]{FFFFED}4.6 & 36.4 & 13.4 & 28.9 & 30.8 & 16.3 & 27.2 \\ 

Final: \textbf{OW-DETR} & \cellcolor[HTML]{FFFFED}\textbf{7.5}  & \textbf{59.2} & \cellcolor[HTML]{FFFFED} \textbf{6.2} & \textbf{53.6} & \textbf{33.5} & \textbf{42.9} & \cellcolor[HTML]{FFFFED} \textbf{5.7} & \textbf{38.3} & \textbf{15.8} & \textbf{30.8} & \textbf{31.4} & \textbf{17.1} & \textbf{27.8} \\
\bottomrule

\end{tabular}%
}\vspace{-0.5em}

\end{table*}

\begin{table}
\centering
\caption{\textbf{Performance comparison on open-set object detection task}. Our OW-DETR generalizes better by effectively modeling the unknowns and decreasing their confusion with known classes.\vspace{-0.2cm}}
\label{tab:open-set}
\setlength{\tabcolsep}{8pt}
\resizebox{0.48\textwidth}{!}{%
\begin{tabular}{@{}l|cc@{}}
\toprule
Evaluated on $\rightarrow$ & Pascal VOC 2007 & Open-Set (WR1) \\ \midrule
Standard Faster R-CNN & 81.8 & 77.1 \\
Standard RetinaNet & 79.2 & 73.8 \\
Dropout Sampling \cite{miller2018dropout} & 78.1 & 71.1 \\
ORE \cite{joseph2021towards} & 81.3 & 78.2 \\
\textbf{Ours: OW-DETR} & \textbf{82.1} & \textbf{78.6} \\ \bottomrule
\end{tabular}%
}\vspace{-0.2cm}

\end{table}

\subsection{Ablation Study\label{sec:ablations}}
Tab.~\ref{tab:ablation_table} shows the impact of progressively integrating our contributions into the baseline framework for the OWOD problem. The comparison is shown in terms of mAP for the known (current and previous) classes and recall for the unknown class, denoted as U-Recall. All the variants shown (except \texttt{Baseline}$^{\dagger}$) include a finetuning step to alleviate the catastrophic forgetting during incremental learning stage. Here, our baseline is the standard Deformable DETR. We also show the upper bound performance of an oracle, \ie, the baseline trained with ground-truth annotations of the unknown class. The \texttt{Baseline} achieves higher performance on the known classes but cannot detect \textit{any} unknown object, since it is trained with only known classes and is thereby not suited for OWOD. 
Integrating the novelty classification branch (denoted by \texttt{Baseline}+\texttt{NC}) and employing the pseudo-unknowns selected by our attention-driven pseudo-labeling mechanism for training the novelty classifier enables the detection of unknown instances. Consequently, such an integration achieves unknown recall rates of $6.0$, $4.6$ and $4.6$ for tasks 1, 2 and 3. Our final framework, OW-DETR, obtained by additionally integrating the objectness branch further improves the retrieval of unknown objects in the OWOD setting, achieving U-Recall of $7.5$, $6.2$ and $5.7$ for the same tasks 1, 2 and 3. These results show the effectiveness of our proposed contributions in the OWOD setting for learning a separation between knowns and unknowns through the novelty classification branch and learning to transfer knowledge from known classes to the unknown through the objectness branch.
\\
\noindent\textbf{Open-set Detection Comparison:} A detector's ability to handle unknown instances in open-set data can be measured by the degree of decrease in its mAP value, compared to its mAP on closed set data. We follow the same evaluation protocol of \cite{miller2018dropout} and report the performance in Tab.~\ref{tab:open-set}. By effectively modeling the unknowns, our OW-DETR achieves promising performance in comparison to existing methods.
\\
\noindent\textbf{Qualitative Analysis:} Fig.~\ref{fig:qual} shows qualitative results on example images from the MS-COCO test set, along with their corresponding attention maps $\bm A$. 
The detections for a known class (in \textcolor{darkyellow}{yellow}) and unknown class (in \textcolor{newpurple}{purple}) obtained from our OW-DETR are also overlayed. 
We observe that unknown objects are detected reasonably well, \eg, \textit{skiis} in top-left image, \textit{tennis racket} in top-right image, \textit{frisbee} in bottom-left image.Although few novel objects are missed (\textit{table} in bottom-right image), these results show that our OW-DETR achieves promising performance in detecting unknown objects in the challenging OWOD setting. Additional results are provided in the appendix.

\section{Relation to Prior Art\label{sec:related_work}}
Several works have investigated the problem of standard object detection~\cite{ren2015faster,redmon2016you,he2017mask,girshick2015fast,lin2017focal,nie2019enriched,cao2020sipmask,pang2019efficient}. These approaches work under a strong assumption that the label space of object categories to be encountered during a model's life-cycle is the same as during its training. 
The advent of transformers for natural language processing \cite{vaswani2017attention,Felix2019context} has inspired studies to investigate related ideas for vision tasks~\cite{dosovitskiy2020image,girdhar2019video,khan2021transformers,naseer2021intriguing}, including standard object detection~\cite{carion2020end,zhu2020deformable}.
Different to standard object detection, incremental object detection approaches~\cite{shmelkov2017incremental,peng2020faster} model newer object classes that are introduced in training incrementally and tackle the issue of catastrophic forgetting.
On the other hand, the works of \cite{dhamija2020overlooked,miller2018dropout,miller2019evaluating,hall2020probabilistic} focus on open-set detection, where new unknown objects encountered during test are to be rejected.
In contrast, the recent work of \cite{joseph2021towards} tackles the challenging open-world object detection (OWOD) problem for detecting both known and unknown objects in addition to incrementally learning new object classes. 
Here, we propose an OWOD approach, OW-DETR, in a transformer-based framework~\cite{zhu2020deformable}, comprising the following novel components: attention-driven pseudo-labeling, novelty classification and objectness scoring. Our OW-DETR explicitly encodes multi-scale contextual information with fewer inductive biases while simultaneously enabling transfer of objectness knowledge from known classes to the novel class for improved unknown detection.

\section{Conclusions\label{sec:conclusion}}
We proposed a novel transformer-based approach, OW-DETR, for the problem of open-world object detection. The proposed OW-DETR comprises dedicated components to address open-world settings, including attention-driven pseudo-labeling, novelty classification and objectness scoring in order to accurately detect unknown objects in images. We conduct extensive experiments on two  popular benchmarks: PASCAL VOC and MS COCO. Our OW-DETR consistently outperforms the recently introduced ORE for all task settings on the MS COCO dataset. Furthermore, OW-DETR achieves state-of-the-art performance in case of incremental object detection on PASCAL VOC dataset.

\section*{Acknowledgements}
This work was partially supported by VR starting grant (2016-05543).



\appendix
\section{Additional Quantitative Results\label{sec:quant}}

\subsection{Evaluation using WI and A-OSE Metrics\label{sec:wi_aose}}
Tab.~\ref{tab:wi_ose} shows a state-of-the-art comparison the for open-world object detection (OWOD) setting on the MS-COCO dataset in terms of wilderness impact (WI) and absolute open-set error (A-OSE). The WI metric \cite{dhamija2020overlooked,joseph2021towards} measures the model's confusion in predicting an unknown instance as known class, given by 
$$ \text{WI} = \frac{P_\mathcal{K}}{P_{\mathcal{K}\cup\mathcal{U}}} - 1, $$
where $P_\mathcal{K}$ is the model precision for known classes when evaluated on known class instances alone and $P_{\mathcal{K}\cup\mathcal{U}}$ denotes the same when evaluated with unknown class instances included. On the other hand, the A-OSE metric measures the total number of unknown instances detected as one of the known classes. Both these two (WI and A-OSE) indicate the degree of confusion in predicting the known classes in the presence of unknown instances. Furthermore, we also show the comparison in terms of U-Recall for ease of comparison. It is worth mentioning that U-Recall directly relates to the unknown class and measures the model's ability to retrieve the unknown instances. 

The standard object detectors (Faster R-CNN and DDETR) in the top part of Tab.~\ref{tab:wi_ose} are inherently not suited for the OWOD setting since they cannot detect any unknown object. Thereby, for these frameworks, only WI and A-OSE can be computed but not U-Recall. Since the energy-based unknown identifier (EBUI) in the recently introduced ORE~\cite{joseph2021towards} is learned using a held-out validation set with weak unknown supervision, for a fair comparison in the OWOD setting, we compare with ORE not employing EBUI. 
We observe that the standard single-stage DDETR wrongly predicts unknown instances as known classes and performs poorly in terms of A-OSE, compared to the two-stage Faster R-CNN. However, by adapting DDETR to OWOD setting through the proposed introduction of attention-driven pseudo-labeling, novelty classification and objectness branch, our OW-DETR obtains improved performance in terms of all three metrics across tasks over the Faster R-CNN based ORE. These results emphasize the importance of the proposed contributions towards a more accurate OWOD.

\subsection{Proposed MS-COCO Split for Open-world\label{sec:proposed_split}}

Open-world object detection (OWOD) is a challenging setting due to its open-taxonomy nature. However, the dataset split proposed in ORE~\cite{joseph2021towards} for OWOD allows data leakage across tasks since different classes from a super-categories are introduced in different tasks, \eg, most classes from \textit{vehicle} and \textit{animal} super-categories are introduced in Task 1, while related classes like \textit{truck}, \textit{elephant}, \textit{bear}, \textit{zebra} and \textit{giraffe} are introduced in Task 2. 
Here, we conduct an experiment by constructing a stricter MS-COCO split, where classes are added across super-categories, as shown in Tab.~\ref{tab:data_split}. Such a split mitigates possible data leakage across tasks since all the classes of a super-category are introduced at a time in a task and not spread across tasks. Thereby, the proposed split is more challenging for OWOD setting. The new split is divided by super-categories with nearly 20 classes in each task: \textit{Animals}, \textit{Person}, \textit{Vehicles} in Task 1; \textit{Appliances}, \textit{Accessories}, \textit{Outdoor}, \textit{Furniture} in Task 2; \textit{Food}, \textit{Sport} in Task 3; \textit{Electronic}, \textit{Indoor}, \textit{Kitchen} in Task 4. Here, Tab.~\ref{tab:new_split_r1} shows that our OW-DETR achieves improved performance even on this stricter OWOD split, compared to the recently introduced ORE. We note that the proposed bottom-up attention driven pseudo-labeling scheme aids our OWOD framework to better generalize to unseen super-categories.

\begin{table*}[t]
\centering
\caption{\textbf{State-of-the-art comparison for open-world object detection (OWOD) on MS-COCO.} The comparison is shown in terms of wilderness impact (WI), absolute open set error (A-OSE) and unknown class recall (U-Recall). The unknown recall (U-Recall) metric quantifies a model's ability to retrieve the unknown object instances. 
The standard object detectors (Faster R-CNN and DDETR) in the top part of table \textit{are inherently not suited for the OWOD setting since they cannot detect any unknown object} and thereby U-Recall cannot be computed for them.
For a fair comparison in the OWOD setting, we compare with the recently introduced ORE~\cite{joseph2021towards} not employing EBUI. Our OW-DETR achieves improved WI and A-OSE over ORE across tasks, thereby indicating lesser confusion in detecting unknown instances as known classes. 
Furthermore, our OW-DETR achieves improved U-Recall over ORE across tasks, indicating our model's ability to better detect the unknown instances. Note that WI, A-OSE and U-Recall cannot be computed in Task 4 (and hence not shown) since all 80 classes are known. See Sec.~\ref{sec:wi_aose} for additional details.  \vspace{-0.3cm}}
\label{tab:wi_ose}
\setlength{\tabcolsep}{10pt}
\adjustbox{width=\textwidth}{
\begin{tabular}{@{}l|ccc|ccc|ccc}
\toprule
 \textbf{Task IDs} ($\rightarrow$)& \multicolumn{3}{c|}{\textbf{Task 1}} & \multicolumn{3}{c|}{\textbf{Task 2}} & \multicolumn{3}{c}{\textbf{Task 3}} \\
\midrule

 & \cellcolor[HTML]{FFFFED}{U-Recall} & \cellcolor[HTML]{EDF6FF}{WI} & \cellcolor[HTML]{EDF6FF}{A-OSE} & \cellcolor[HTML]{FFFFED}{U-Recall} & \cellcolor[HTML]{EDF6FF}{WI} & \cellcolor[HTML]{EDF6FF}{A-OSE}  & \cellcolor[HTML]{FFFFED}{U-Recall} & \cellcolor[HTML]{EDF6FF}{WI} & \cellcolor[HTML]{EDF6FF}{A-OSE} \\

 & \cellcolor[HTML]{FFFFED}($\uparrow$) & \cellcolor[HTML]{EDF6FF}($\downarrow$) & \cellcolor[HTML]{EDF6FF}($\downarrow$) & \cellcolor[HTML]{FFFFED}($\uparrow$) & \cellcolor[HTML]{EDF6FF}($\downarrow$) & \cellcolor[HTML]{EDF6FF}($\downarrow$) & \cellcolor[HTML]{FFFFED}($\uparrow$) & \cellcolor[HTML]{EDF6FF}($\downarrow$) & \cellcolor[HTML]{EDF6FF}($\downarrow$) \\

 \midrule

Faster-RCNN~\cite{ren2015faster}  & -  & 0.0699 & 13396 & - & 0.0371 & 12291 & - & 0.0213 & 9174  \\ 
\begin{tabular}[c]{@{}l@{}}Faster-RCNN\\ \hspace{2em}+ Finetuning\end{tabular} & \multicolumn{3}{c|}{\begin{tabular}[c]{@{}c}Not applicable in Task 1\end{tabular}}  & -  & 0.0375 & 12497 & - & 0.0279 & 9622   \\ 


DDETR~\cite{zhu2020deformable}  & -  & 0.0608 & 33270 & - & 0.0368 & 18115 & - & 0.0197 & 9392  \\ 

\begin{tabular}[c]{@{}l@{}}DDETR\\ \hspace{2em}+ Finetuning\end{tabular} & \multicolumn{3}{c|}{\begin{tabular}[c]{@{}c}Not applicable in Task 1\end{tabular}} & -  & 0.0337 & 17834 & - & 0.0195 & 10095 \\ 
\midrule
\midrule 

ORE $-$ EBUI ~\cite{joseph2021towards} & \cellcolor[HTML]{FFFFED}4.9  & 0.0621 & 10459 & \cellcolor[HTML]{FFFFED}2.9 & 0.0282 & 10445 & \cellcolor[HTML]{FFFFED}3.9 & 0.0211 & 7990  \\ 

\textbf{Ours: OW-DETR} & \cellcolor[HTML]{FFFFED}\textbf{7.5}  & \textbf{0.0571} & \textbf{10240} & \cellcolor[HTML]{FFFFED}\textbf{6.2} & \textbf{0.0278} & \textbf{8441} & \cellcolor[HTML]{FFFFED}\textbf{5.7} & \textbf{0.0156} & \textbf{6803}  \\
\bottomrule

\end{tabular}%
}

\end{table*}

\begin{table*}[t]
\centering
\caption{\textbf{State-of-the-art comparison for OWOD on the proposed MS-COCO split.} The comparison is shown in terms of known class mAP and unknown class recall (U-Recall). For a fair comparison in the OWOD setting, we compare with the recently introduced ORE~\cite{joseph2021towards} not employing EBUI. The proposed split mitigates data leakage across tasks and is more challenging than the original OWOD split of~\cite{joseph2021towards}. Even on this harder data split, our OW-DETR achieves improved U-Recall over ORE across tasks, indicating our model's ability to better detect the unknown instances. Furthermore, our OW-DETR also achieves significant gains in mAP for the known classes across the four tasks. Note that since all 80 classes are known in Task 4, U-Recall is not computed. See Sec.~\ref{sec:proposed_split} for more details.  \vspace{-0.3cm}}

\label{tab:new_split_r1}
\setlength{\tabcolsep}{2pt}
\adjustbox{width=\textwidth}{
\begin{tabular}{@{}l|cc|cccc|cccc|ccc@{}}
\toprule
 \textbf{Task IDs} ($\rightarrow$)& \multicolumn{2}{c|}{\textbf{Task 1}} & \multicolumn{4}{c|}{\textbf{Task 2}} & \multicolumn{4}{c|}{\textbf{Task 3}} & \multicolumn{3}{c}{\textbf{Task 4}} \\ \midrule
 
 & \cellcolor[HTML]{FFFFED}{U-Recall} & \multicolumn{1}{c|}{\cellcolor[HTML]{EDF6FF}{mAP ($\uparrow$)}} & \cellcolor[HTML]{FFFFED}{U-Recall} & \multicolumn{3}{c|}{\cellcolor[HTML]{EDF6FF}{mAP ($\uparrow$)}} & \cellcolor[HTML]{FFFFED}{U-Recall} & \multicolumn{3}{c|}{\cellcolor[HTML]{EDF6FF}{mAP ($\uparrow$)}} & \multicolumn{3}{c}{\cellcolor[HTML]{EDF6FF}{mAP ($\uparrow$)}}  \\

 & \cellcolor[HTML]{FFFFED}($\uparrow$) & \begin{tabular}[c]{@{}c}Current \\ known\end{tabular} & \cellcolor[HTML]{FFFFED}($\uparrow$) & \begin{tabular}[c]{@{}c@{}}Previously\\  known\end{tabular} & \begin{tabular}[c]{@{}c@{}}Current \\ known\end{tabular} & Both & \cellcolor[HTML]{FFFFED}($\uparrow$) & \begin{tabular}[c]{@{}c@{}}Previously \\ known\end{tabular} & \begin{tabular}[c]{@{}c@{}}Current \\ known\end{tabular} & Both & \begin{tabular}[c]{@{}c@{}}Previously \\ known\end{tabular} & \begin{tabular}[c]{@{}c@{}}Current \\ known\end{tabular} & Both \\ \midrule

ORE $-$ EBUI ~\cite{joseph2021towards} & \cellcolor[HTML]{FFFFED}1.5  & 61.4 & \cellcolor[HTML]{FFFFED}3.9 & 56.5 & 26.1 & 40.6  & \cellcolor[HTML]{FFFFED}3.6 & 38.7 & 23.7 & 33.7 & 33.6 & 26.3 & 31.8 \\ 

\textbf{Ours: OW-DETR} & \cellcolor[HTML]{FFFFED}\textbf{5.7}  & \textbf{71.5} & \cellcolor[HTML]{FFFFED}\textbf{6.2} & \textbf{62.8} & \textbf{27.5} & \textbf{43.8} & \cellcolor[HTML]{FFFFED}\textbf{6.9} & \textbf{45.2} & \textbf{24.9} & \textbf{38.5} & \textbf{38.2} & \textbf{28.1} & \textbf{33.1} \\
\bottomrule

\end{tabular}%
}

\end{table*}

\begin{table}
\centering
\caption{\textbf{Task composition in the proposed MS-COCO split for Open-world evaluation protocol}. The semantics of each task and the number of images and instances (objects) across splits are shown. The proposed task split mitigates the data leakage across tasks that was present in the split of ORE~\cite{joseph2021towards}. \Eg, all \textit{vehicles} including \textit{truck}, which was in Task 2 earlier are now in Task 1. Similarly all \textit{animals} are now in Task 1, while other Pascal VOC classes like \textit{sofa}, \textit{bottle}, \etc are moved out of Task 1.\vspace{-0.2cm}}
\resizebox{\columnwidth}{!}{%
\begin{tabular}{@{}l|cccc@{}}
\toprule
 & \textbf{Task 1} & \textbf{Task 2} & \textbf{Task 3} & \textbf{Task 4} \\ \midrule
Semantic split & \begin{tabular}[c]{@{}c@{}}Animals, Person, \\ Vehicles\end{tabular} & \begin{tabular}[c]{@{}c@{}}Appliances, Accessories, \\ Outdoor, Furniture\end{tabular} & \begin{tabular}[c]{@{}c@{}}Sports, \\ Food\end{tabular} & \begin{tabular}[c]{@{}c@{}}Electronic, Indoor, \\ Kitchen\end{tabular} \\\midrule
\# training images & 89490 & 55870 & 39402 & 38903 \\
\# test images & 3793 & 2351 & 1642 & 1691 \\
\# train instances & 421243 & 163512 & 114452 & 160794 \\
\# test instances & 17786 & 7159 & 4826 & 7010 \\ \bottomrule
\end{tabular}%
}\vspace{-0.5em}

\label{tab:data_split}
\end{table}

\subsection{Fully- \vs Self-supervised Pretraining\label{sec:fs_dino}}
As discussed in the implementation details, our OW-DETR framework employs a ResNet-50 backbone that is pretrained on ImageNet1K~\cite{imagenet} in a self-supervised manner~\cite{caron2021emerging} (DINO) without labels. Such a pretraining mitigates a likely open-world setting violation, which could occur in fully-supervised (FS) pretraining, with class labels, due to possible overlap with the novel classes. Here, we additionally evaluate the performance of employing the ResNet-50 backbone, which is pretrained in an FS manner. Tab.~\ref{tab:fullysuper_resnet} shows the performance comparison between FS and DINO pretraining of ResNet-50. We observe that DINO pretraining enables a stronger backbone and achieves improved performance over FS pretraining for OWOD while additionally mitigating the violation in open-world setting.

\begin{table}[t]
\centering
\caption{Comparison of OW-DETR when using ImageNet1K pretrained ResNet-50 trained in (i) fully-supervised (FS) setting using class labels and (ii) self-supervised (DINO) setting without class labels. Note that the FS backbone violates OWOD settings due to overlap between pretraining (\textit{annotated}) classes and unknowns. Hence, we utilize DINO ResNet50 for a fair OWOD evaluation. 
 \vspace{-0.3cm}}
\label{tab:fullysuper_resnet}
\setlength{\tabcolsep}{5pt}
\adjustbox{width=\columnwidth}{
\begin{tabular}{@{}l|cc|cc|cc|c}
\toprule
 \textbf{Backbone} & \multicolumn{2}{c|}{\textbf{Task 1}} & \multicolumn{2}{c|}{\textbf{Task 2}} & \multicolumn{2}{c|}{\textbf{Task 3}} & \multicolumn{1}{c}{\textbf{Task 4}} \\ 
 
  & \cellcolor[HTML]{FFFFED}{U-Recall} & \multicolumn{1}{c|}{\cellcolor[HTML]{EDF6FF}{ mAP }} & \cellcolor[HTML]{FFFFED}{U-Recall} & \multicolumn{1}{c|}{\cellcolor[HTML]{EDF6FF}{ mAP }} & \cellcolor[HTML]{FFFFED}{U-Recall} & \multicolumn{1}{c|}{\cellcolor[HTML]{EDF6FF}{ mAP }} &  \multicolumn{1}{c}{\cellcolor[HTML]{EDF6FF}{ mAP }} \\ 
\midrule

FS  & \cellcolor[HTML]{FFFFED}6.2  & 57.6 & \cellcolor[HTML]{FFFFED}5.6 & 40.2 & \cellcolor[HTML]{FFFFED} 4.1 & 30.0  & 27.2 \\

DINO & \cellcolor[HTML]{FFFFED}\textbf{7.5}  & \textbf{59.2} & \cellcolor[HTML]{FFFFED}\textbf{6.2} & \textbf{42.9} & \cellcolor[HTML]{FFFFED} \textbf{5.7} & \textbf{30.8} & \textbf{27.8} \\

\bottomrule

\end{tabular}%
}\vspace{-0.25cm}

\end{table}

\section{Additional Qualitative Results\label{sec:qual}}

\begin{figure*}[t]
  \centering\setlength{\tabcolsep}{3pt}
  \resizebox{\textwidth}{!}{%
    \begin{tabular}{c|c}
        \includegraphics[width=0.492\textwidth]{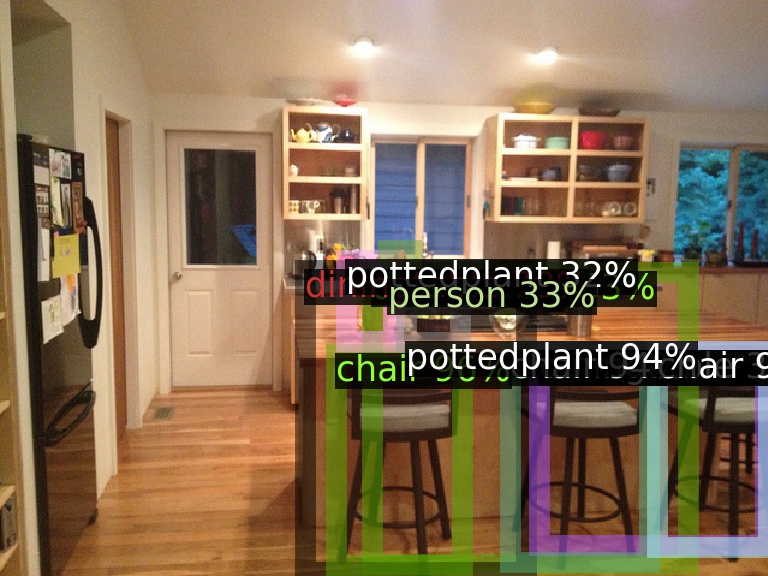} &
      \includegraphics[width=0.5\textwidth, height=0.37\textwidth]{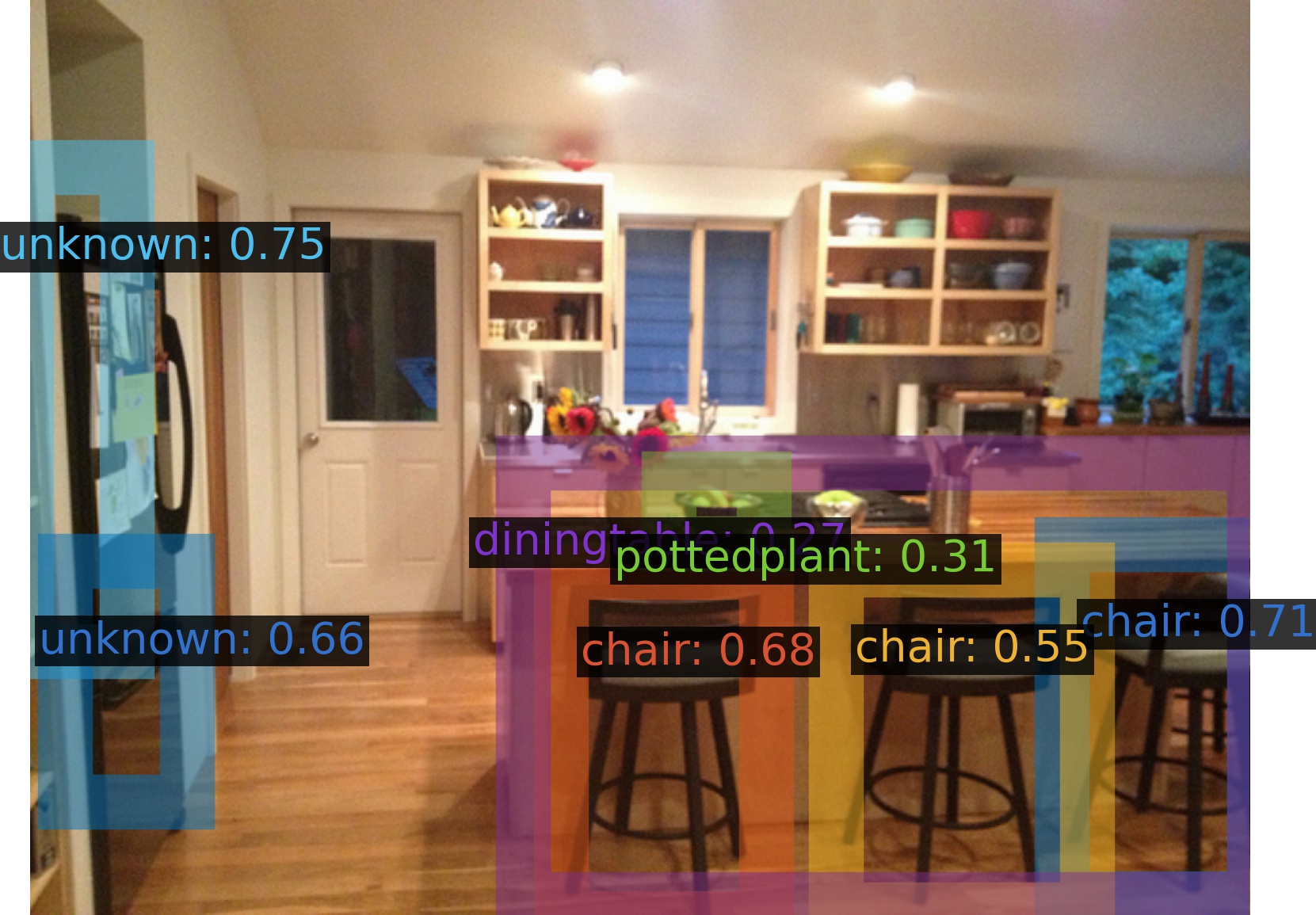}  \\
       \includegraphics[width=0.5\textwidth]{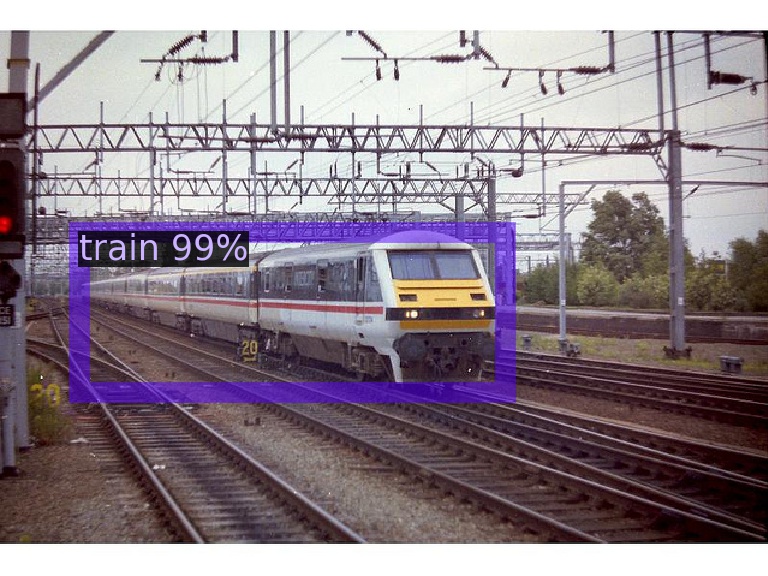} &
      \includegraphics[width=0.5\textwidth]{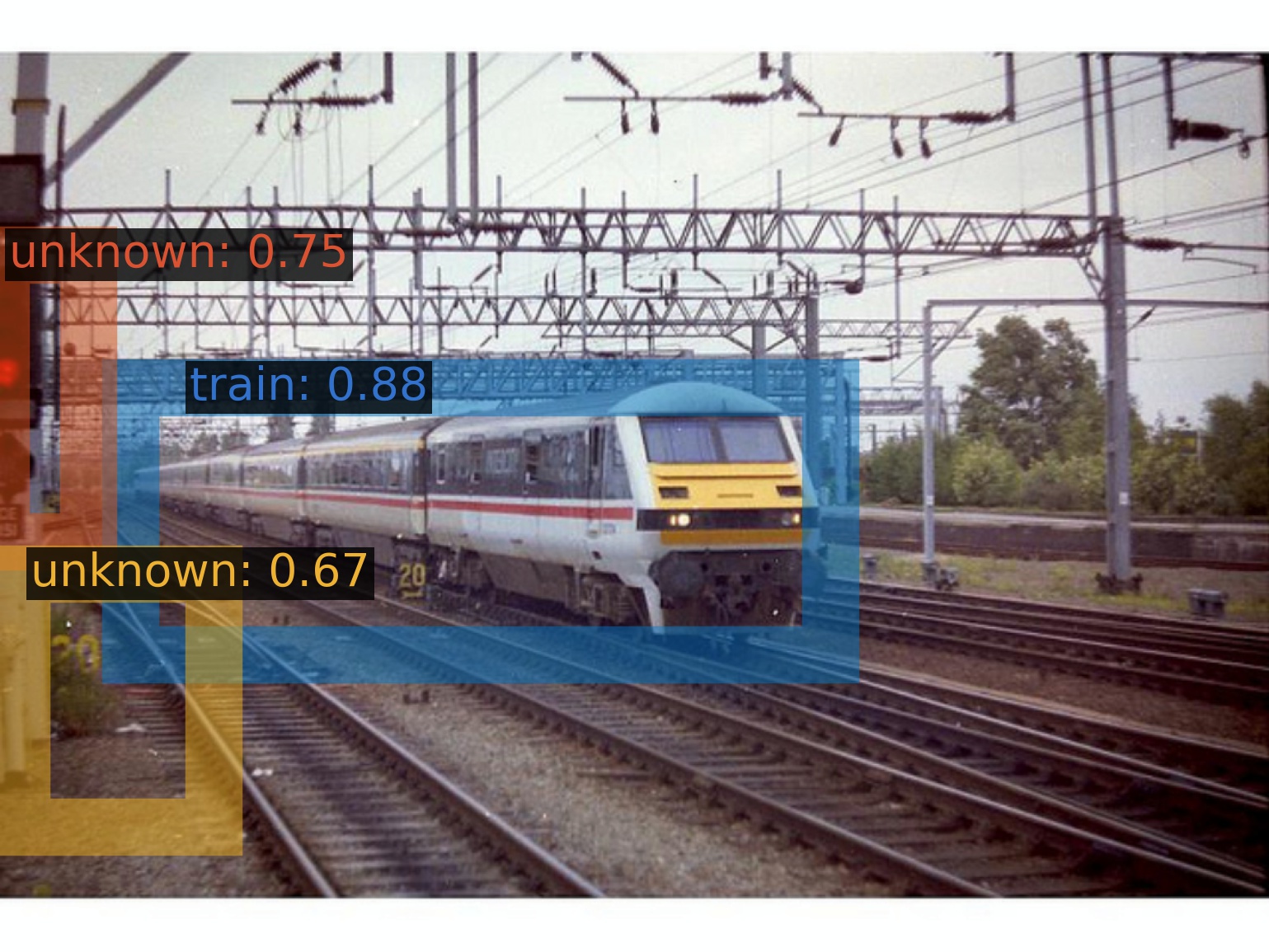} \\
      \includegraphics[width=0.5\textwidth]{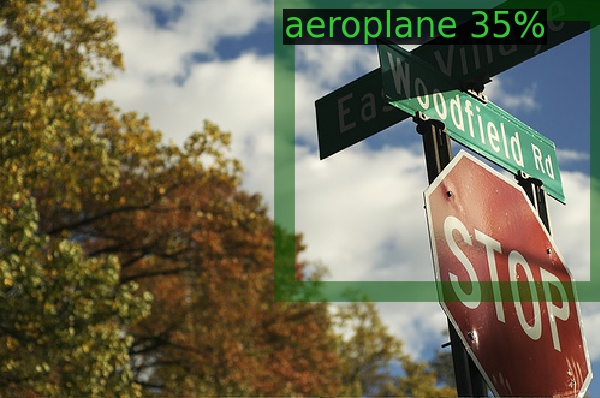} &
      \includegraphics[width=0.5\textwidth]{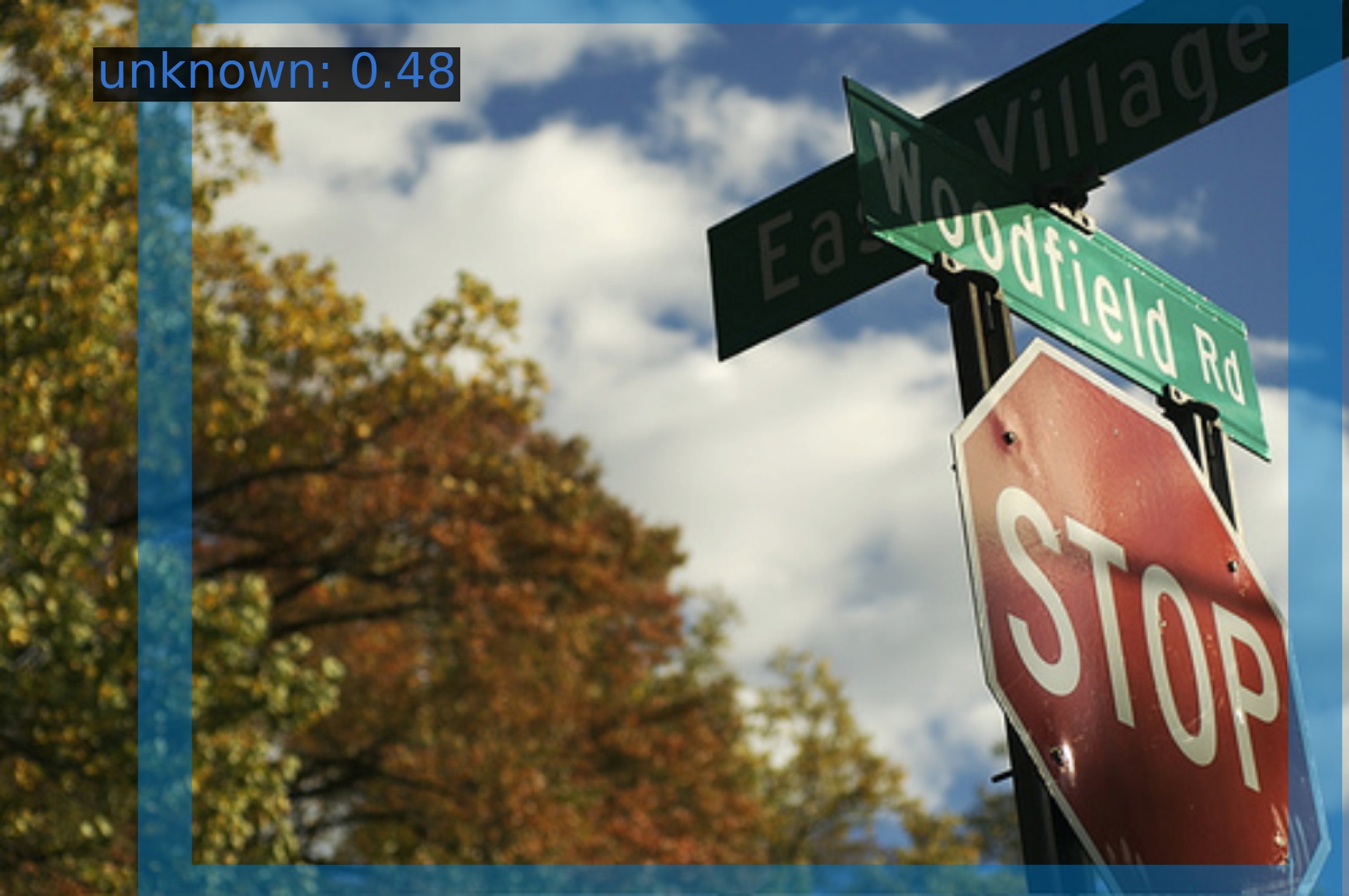} \\
        \textbf{ORE}~\cite{joseph2021towards} & \textbf{Ours: OW-DETR} 
    \end{tabular}}
\vspace{-0.25cm}
\caption{\textbf{OWOD qualitative comparison between ORE~\cite{joseph2021towards} and our OW-DETR on example images in the MS-COCO test-set for Task 1 evaluation.} The predictions of ORE are shown on the left, while those from our OW-DETR are shown on the right. We observe that, in comparison to ORE, our OW-DETR obtains improved detections for the unknown instances. \Eg, in top row, the \textit{refrigerator} (unknown in Task 1) in the left part of the image is detected as unknown by OW-DETR, while it is missed by ORE. Similarly, in the second row, \textit{traffic light} (not part of known classes in Task 1) in the left part of the image are detected by our OW-DETR. Furthermore, while ORE wrongly detects the sign boards as an \textit{aeroplane} in the third row, our OW-DETR detects an unknown object in its place. See Fig.~\ref{fig:qual_t2} for more examples. These results show that the proposed OW-DETR achieves improved detection of unknown objects, in comparison to ORE.}

\label{fig:qual_t1}
\end{figure*}

\begin{figure*}[t]
  \centering\setlength{\tabcolsep}{3pt}
\scalebox{0.9}{

    \begin{tabular}{c|c}
      \includegraphics[width=0.5\textwidth]{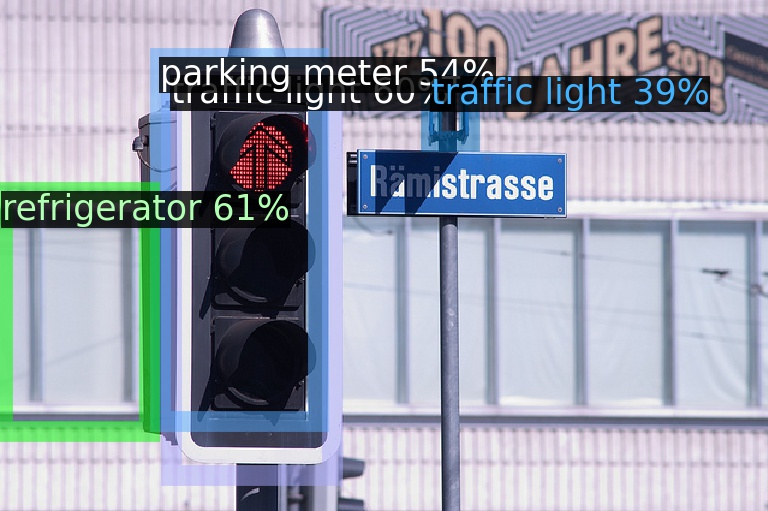} &
      \includegraphics[width=0.5\textwidth]{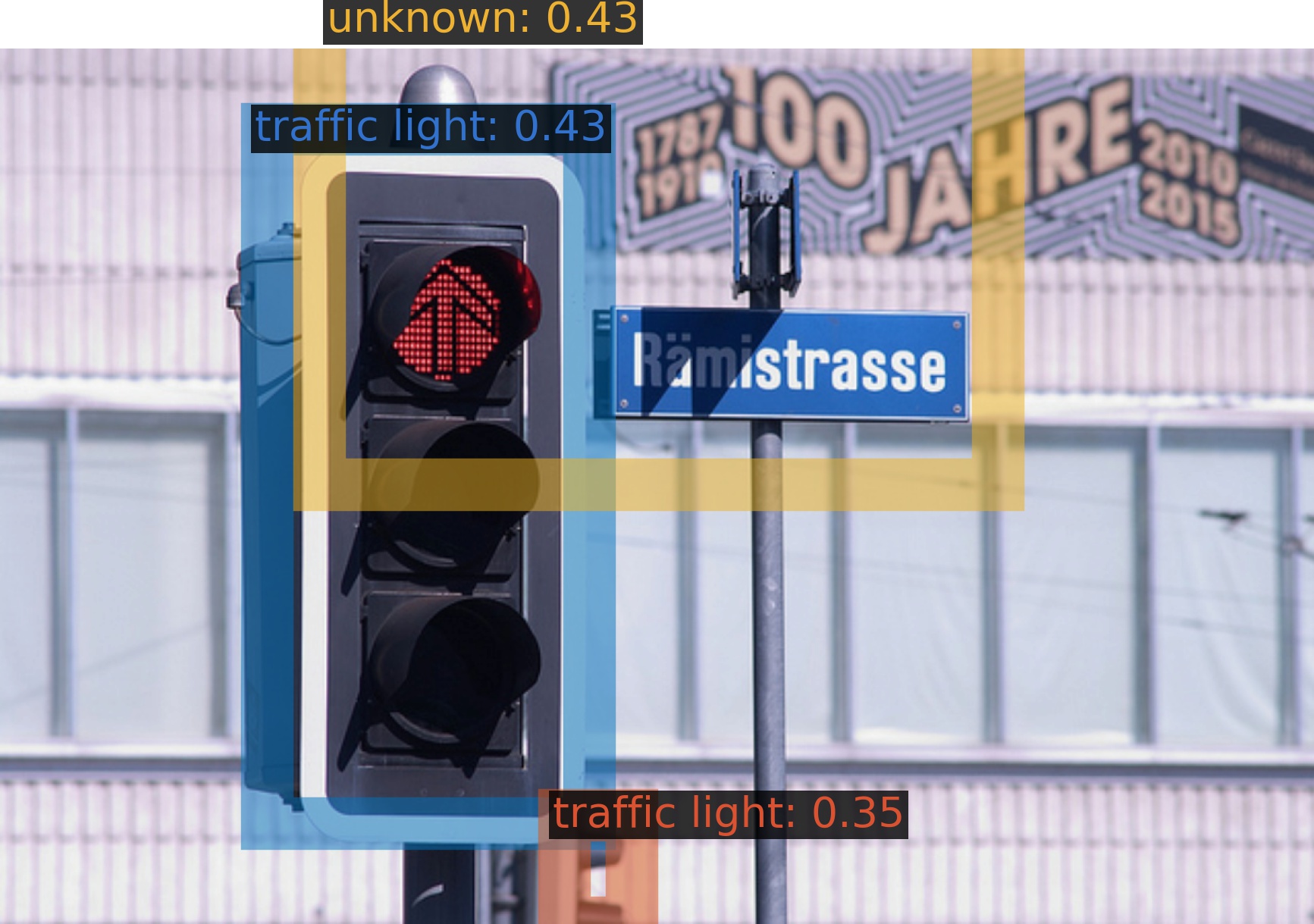}  \\
      \includegraphics[width=0.5\textwidth]{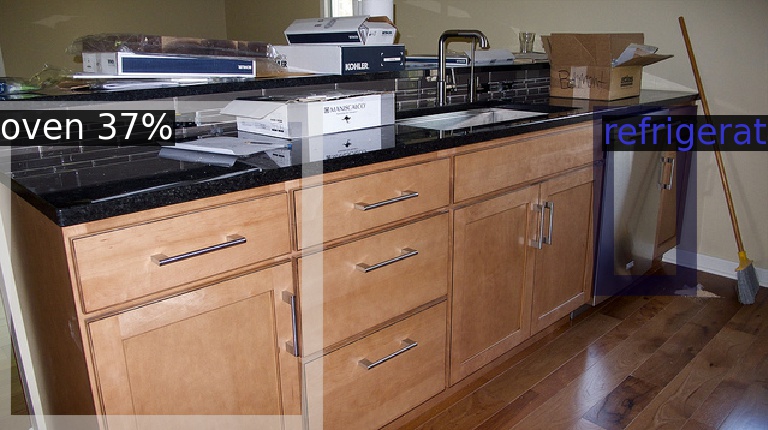} &
      \includegraphics[width=0.5\textwidth]{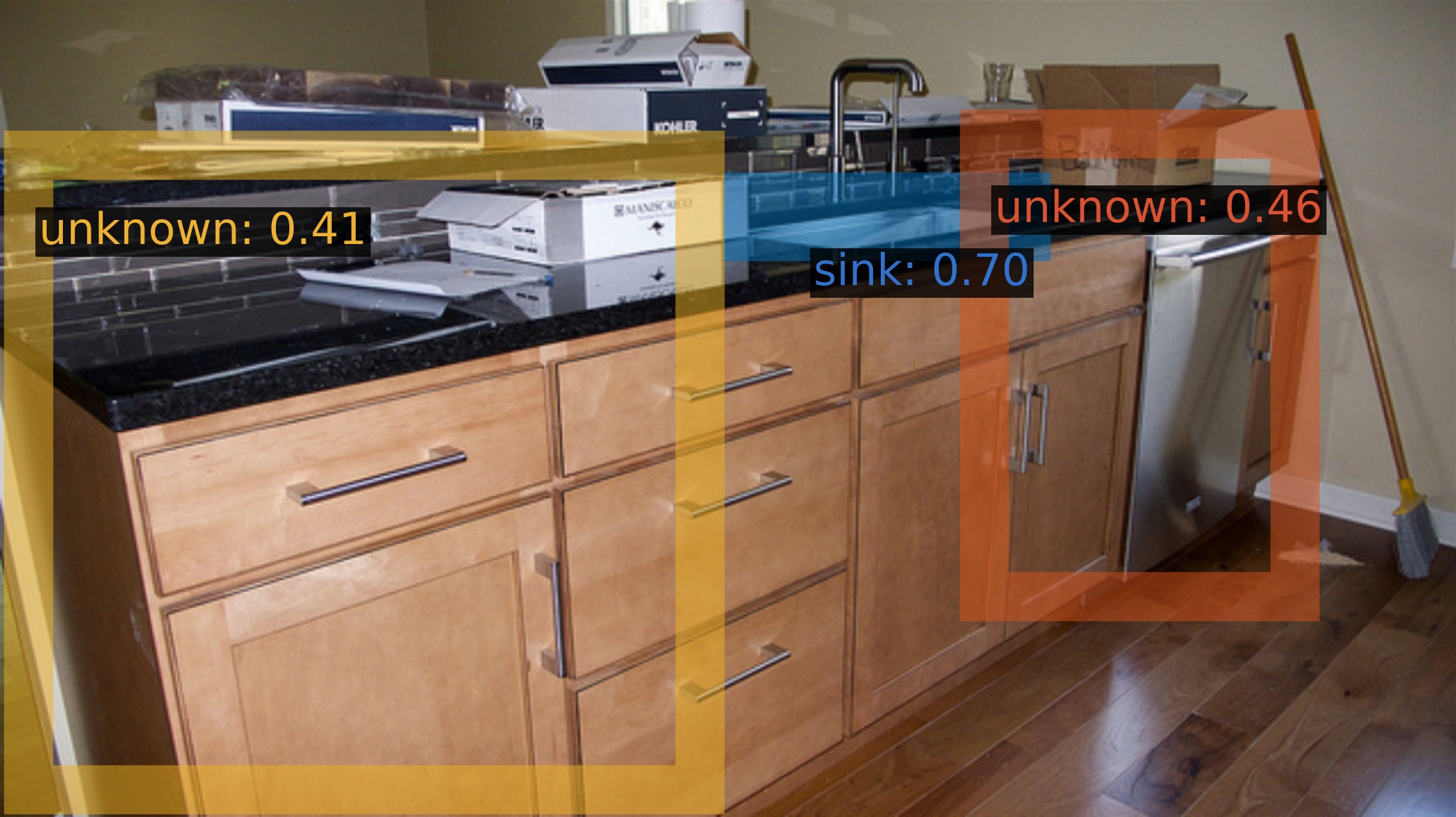}  \\
      \includegraphics[width=0.37\textwidth]{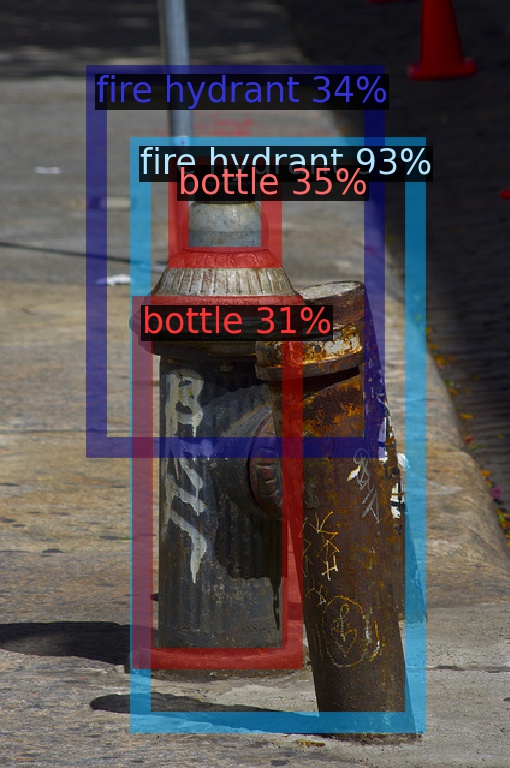} &
      \includegraphics[width=0.37\textwidth]{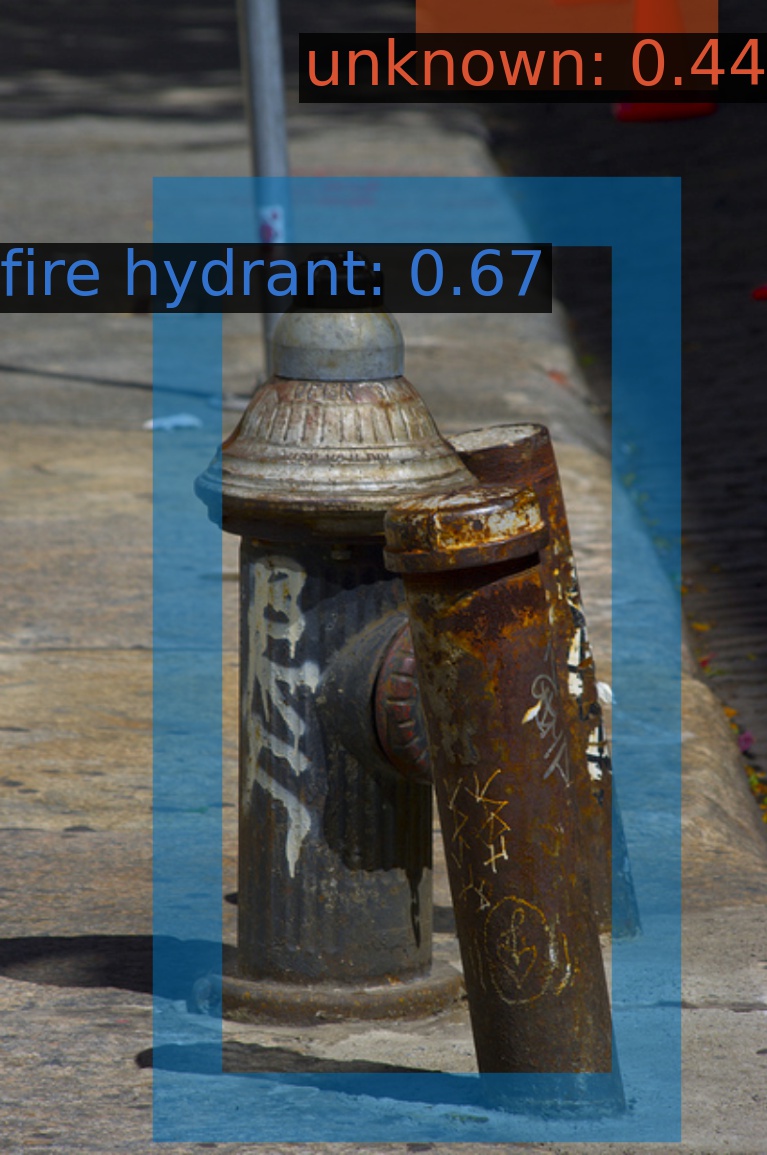} \\
      \textbf{ORE}~\cite{joseph2021towards} & \textbf{Ours: OW-DETR} 

    \end{tabular}
    
    }

\vspace{-0.25cm}
\caption{\textbf{OWOD qualitative comparison between ORE~\cite{joseph2021towards} and our OW-DETR on example images in the MS-COCO test-set for Task 2 evaluation.} The predictions of ORE are shown on the left, while those from the proposed OW-DETR are shown on the right. We observe that, in comparison to ORE, our OW-DETR achieves promising detections for the unknown objects. \Eg, in top row, ORE wrongly predicts \textit{traffic light} on a road sign (true unknown), whereas our OW-DETR correctly detects it as an unknown object. In addition, our OW-DETR also detects the smaller \textit{traffic light} accurately. In the second row, while ORE detects cupboards as \textit{oven}, our OW-DETR detects it as unknown. Furthermore, ORE detects multiple objects on \textit{fire hydrant}, which is mitigated by our OW-DETR. These results show that the proposed OW-DETR captures better reasoning \wrt unknown objects, in comparison to ORE. See Sec.~\ref{sec:qual} for additional details.}

\label{fig:qual_t2}
\end{figure*}

\begin{figure*}[t]
  \centering\setlength{\tabcolsep}{3pt}
  \resizebox{\textwidth}{!}{%
    \begin{tabular}{c|c}
     \includegraphics[width=0.495\textwidth]{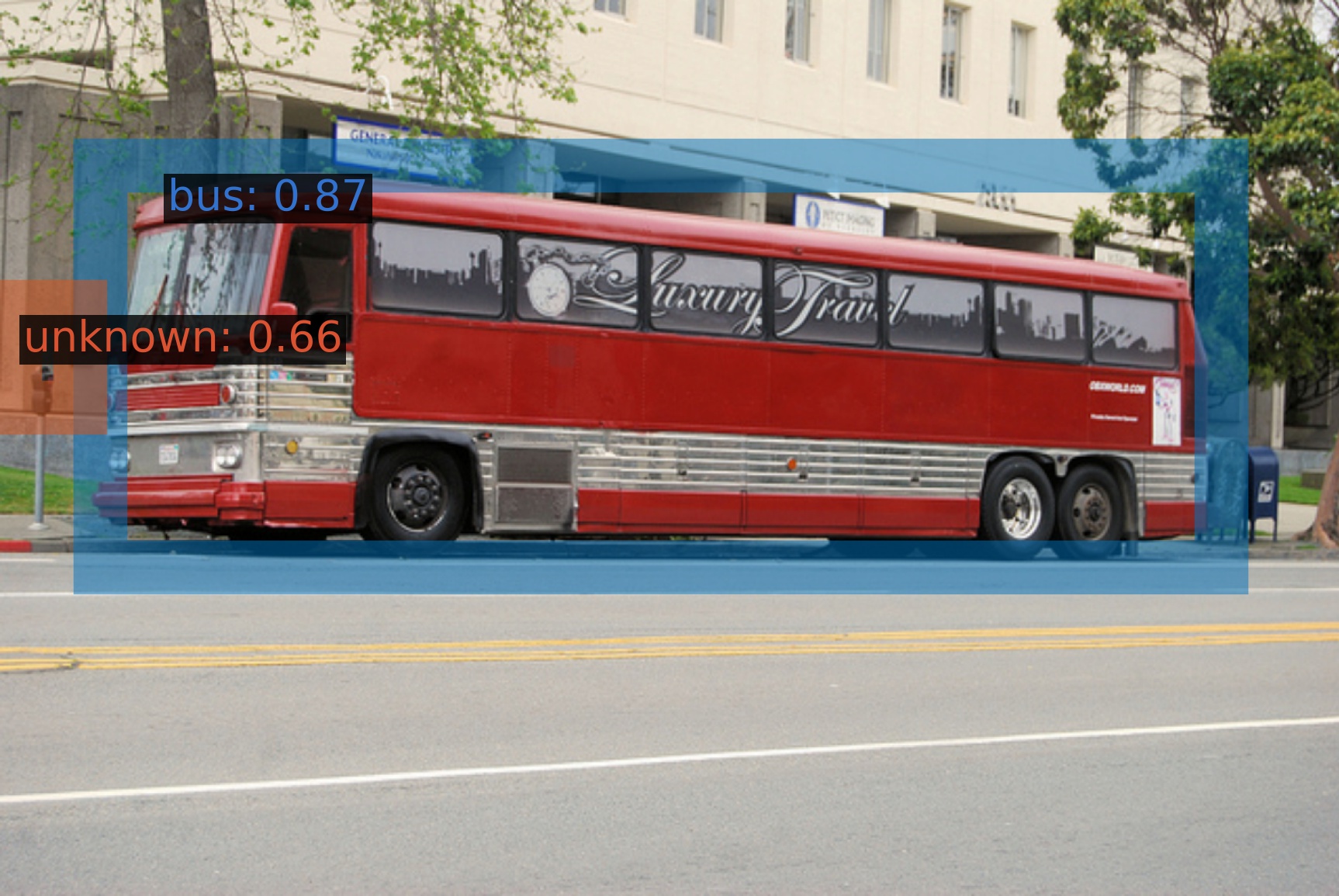}&
      \includegraphics[width=0.5\textwidth]{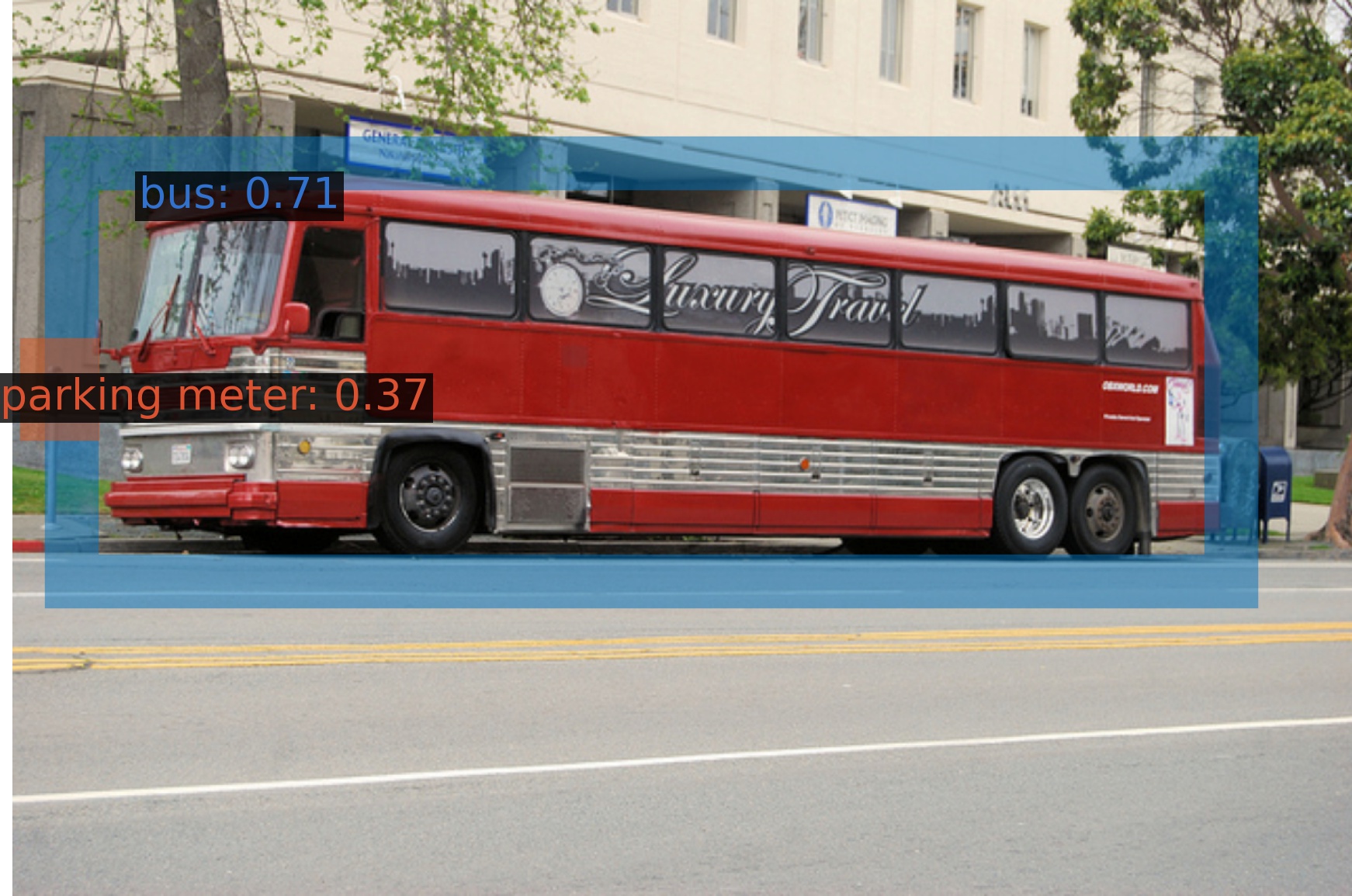} \\
      \includegraphics[width=0.5\textwidth]{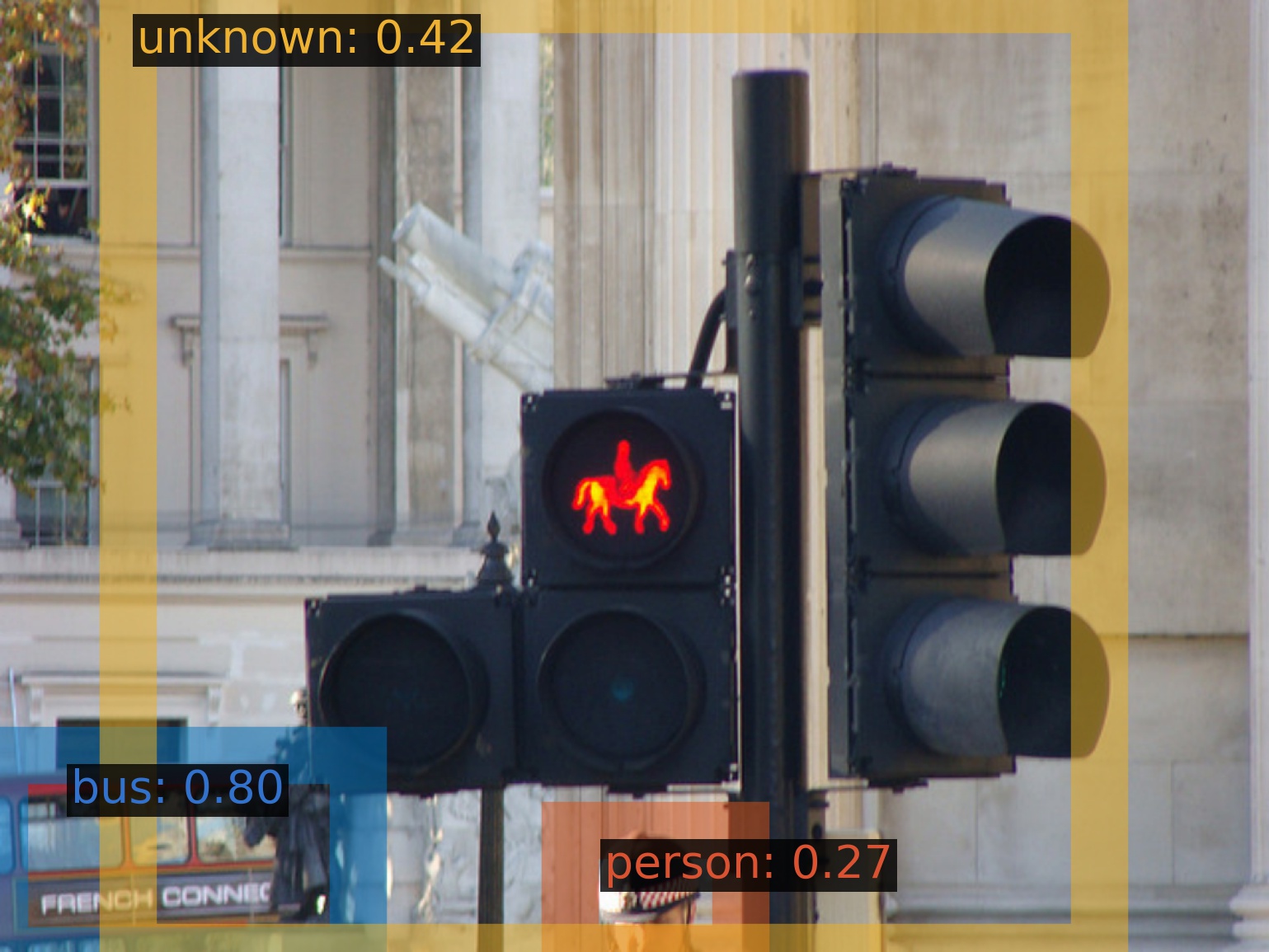}&
      \includegraphics[width=0.5\textwidth]{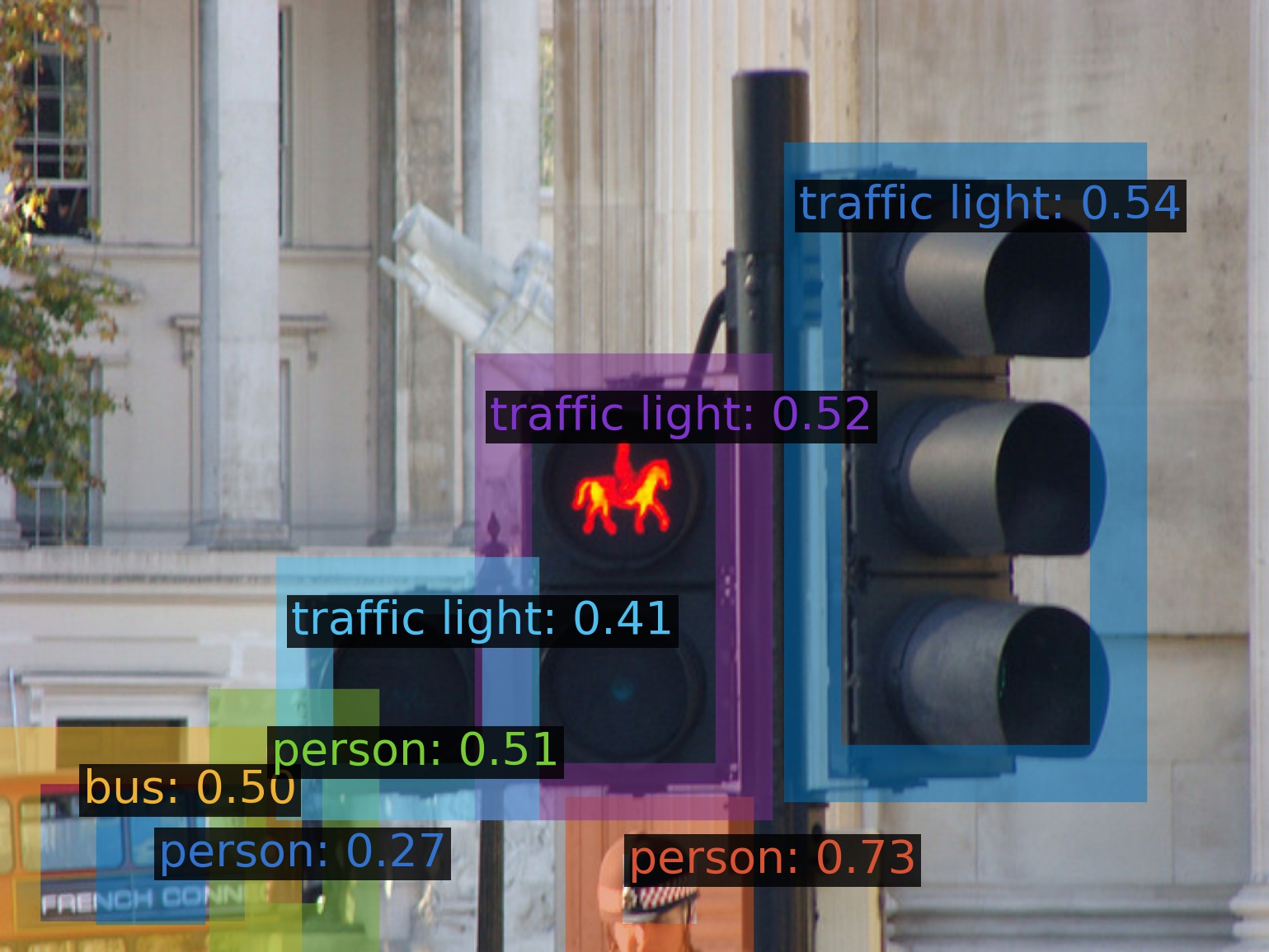}  \\
      \includegraphics[width=0.5\textwidth]{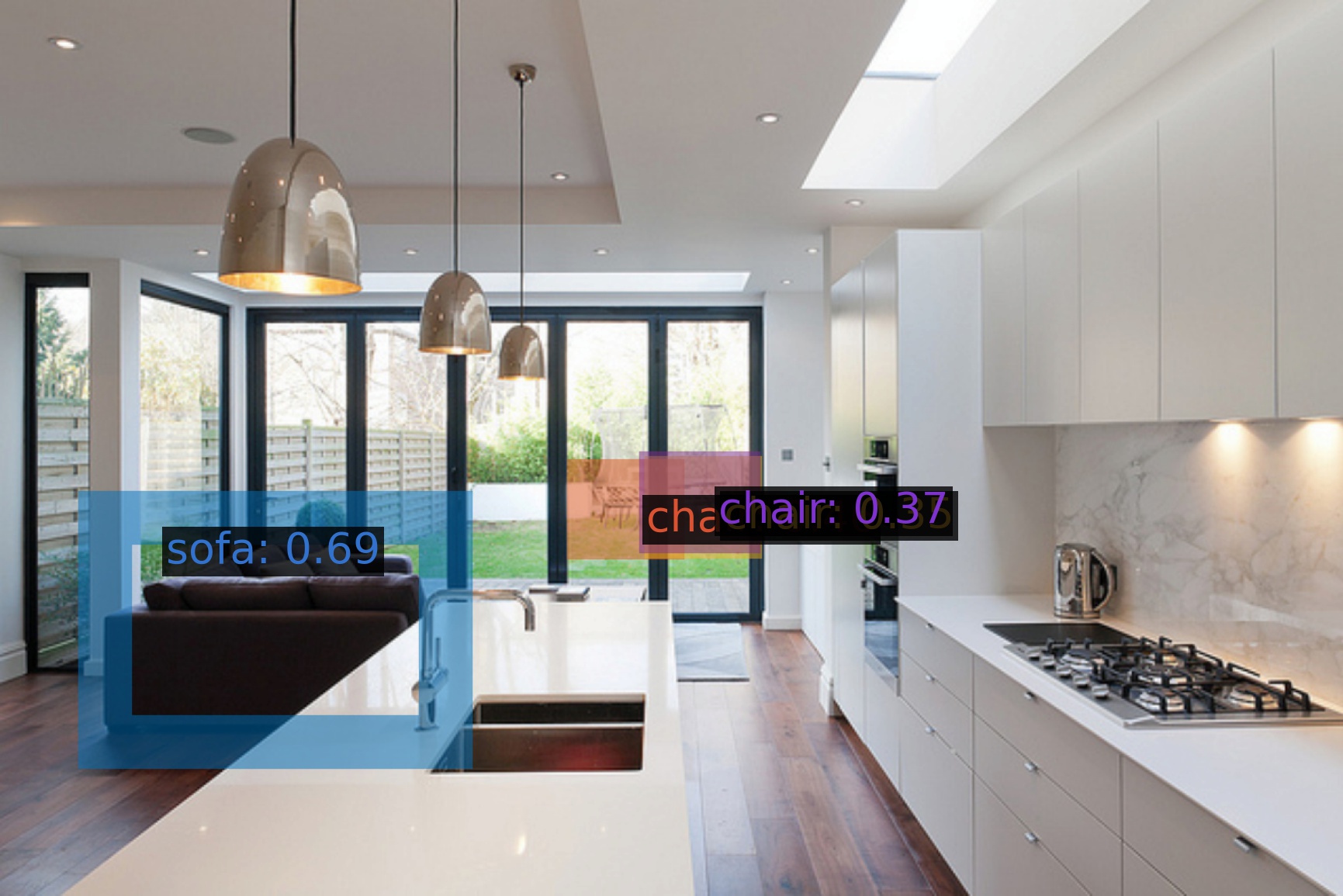}&
      \includegraphics[width=0.5\textwidth]{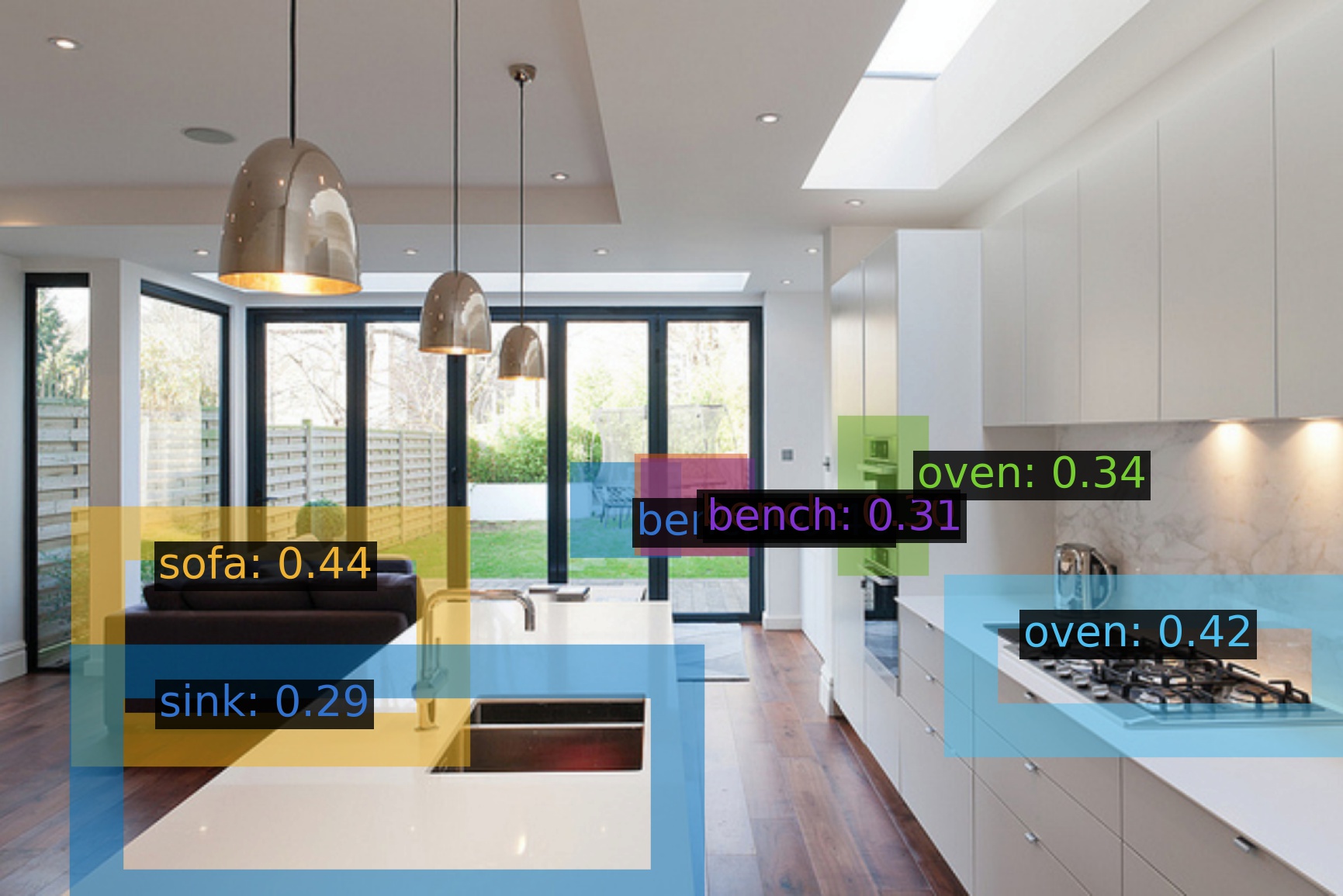}  \\
      \textbf{Task 1 evaluation} & \textbf{Task 2 evaluation} 
      
    \end{tabular}}
\vspace{-0.25cm}
\caption{\textbf{Illustration showing the evolution of predictions of the proposed OW-DETR in the OWOD setting on MS-COCO images.} The objects detected by our OW-DETR when trained only on Task-1 classes is shown on the left. The predictions for the same images after incrementally training with Task 2 classes is shown on the right. In the top row, an unknown prediction during Task 1 evaluation is correctly predicted as \textit{parking meter} during Task 2 evaluation. In the second row, \textit{traffic lights} that are correctly detected as unknown objects during Task 1 evaluation are correctly detected as known objects during Task 2 evaluation. In the third row, potential unknown objects (\textit{bench}) are detected but confused as \textit{chair} due to their visual similarity during Task 1. However, they are correctly classified in Task 2 after \textit{bench} class is incrementally learned. These results show promising performance of our OW-DETR in initially detecting potential unknown objects and later correctly detecting them when their corresponding classes are incrementally introduced for learning.}
\label{fig:incr_1}
\end{figure*}

\begin{figure*}[t]
  \centering\setlength{\tabcolsep}{3pt}
  \resizebox{\textwidth}{!}{%
    \begin{tabular}{c|c}
 
      \includegraphics[width=0.5\textwidth]{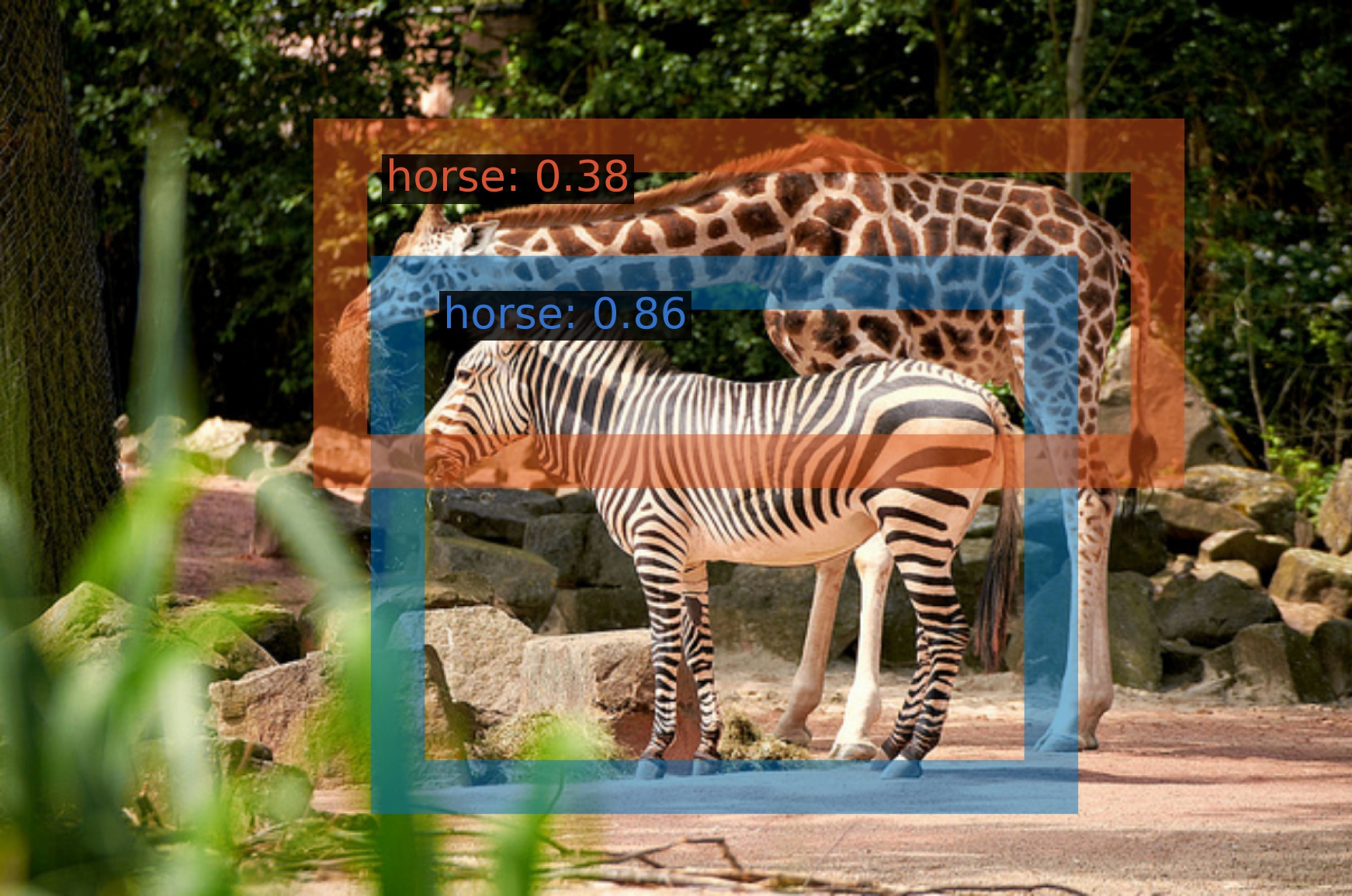} & \hspace{-1.2em}
      \includegraphics[width=0.5\textwidth]{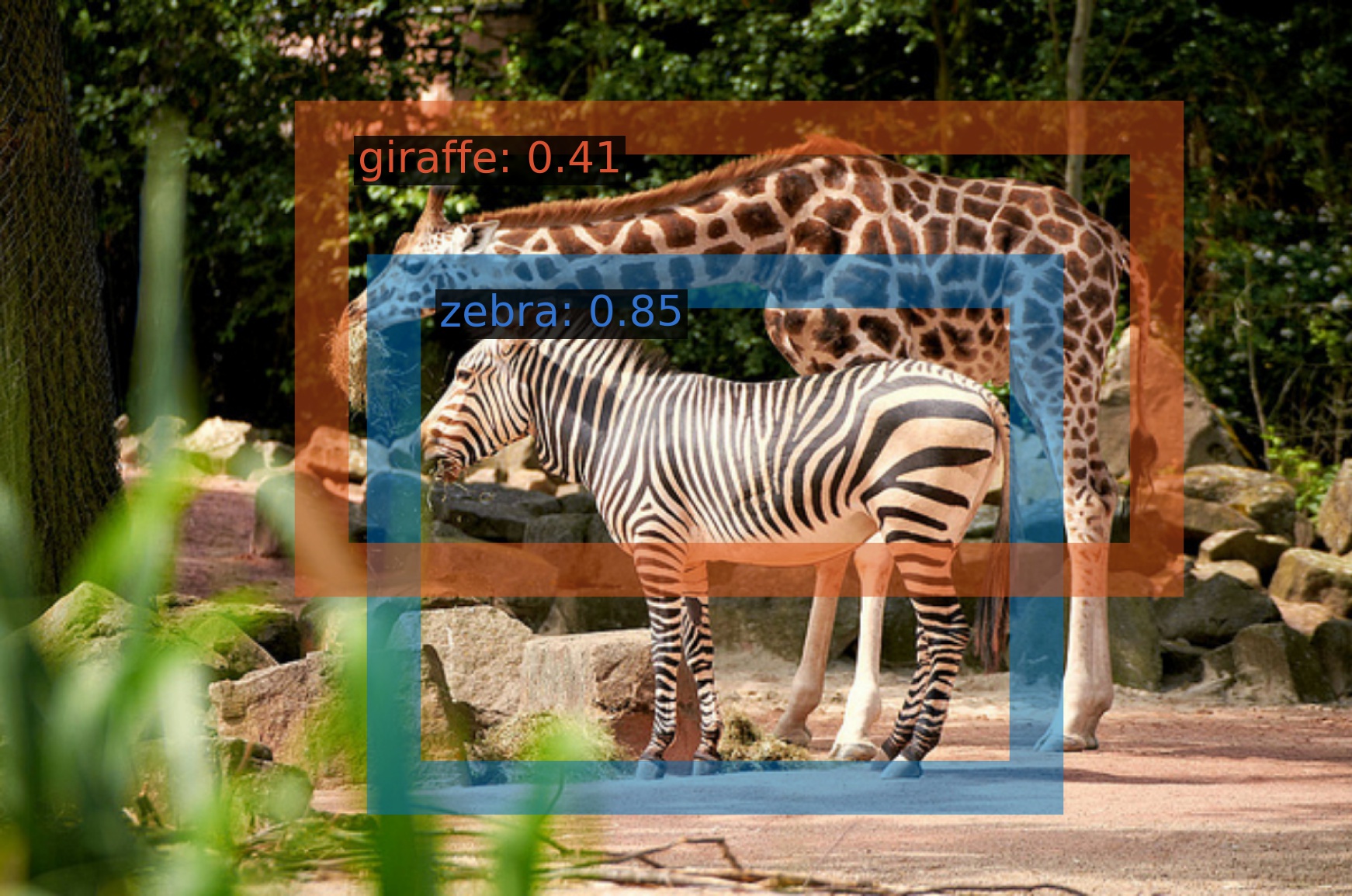}  \\ 
      \includegraphics[width=0.5\textwidth]{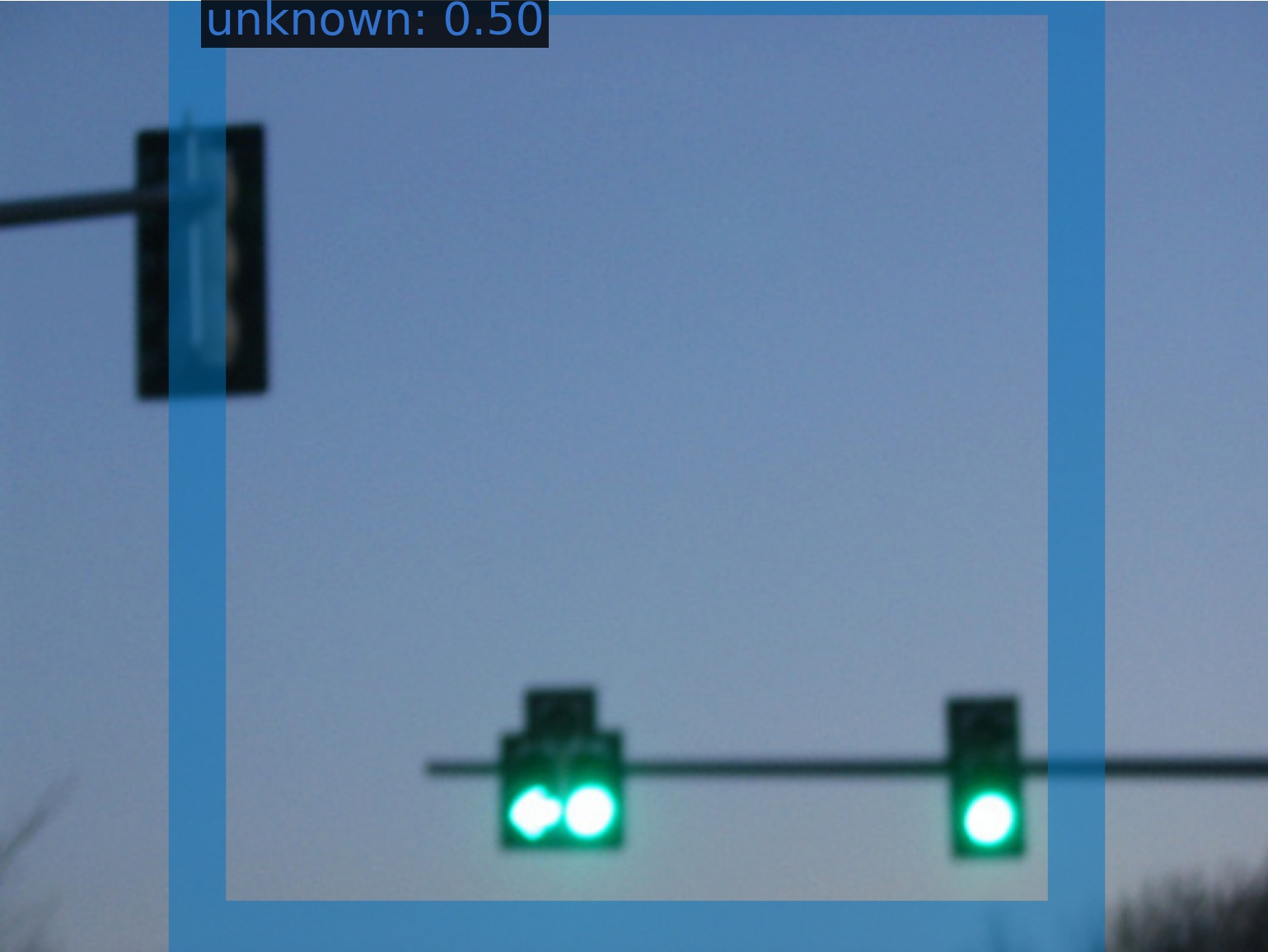}&
      \includegraphics[width=0.52\textwidth]{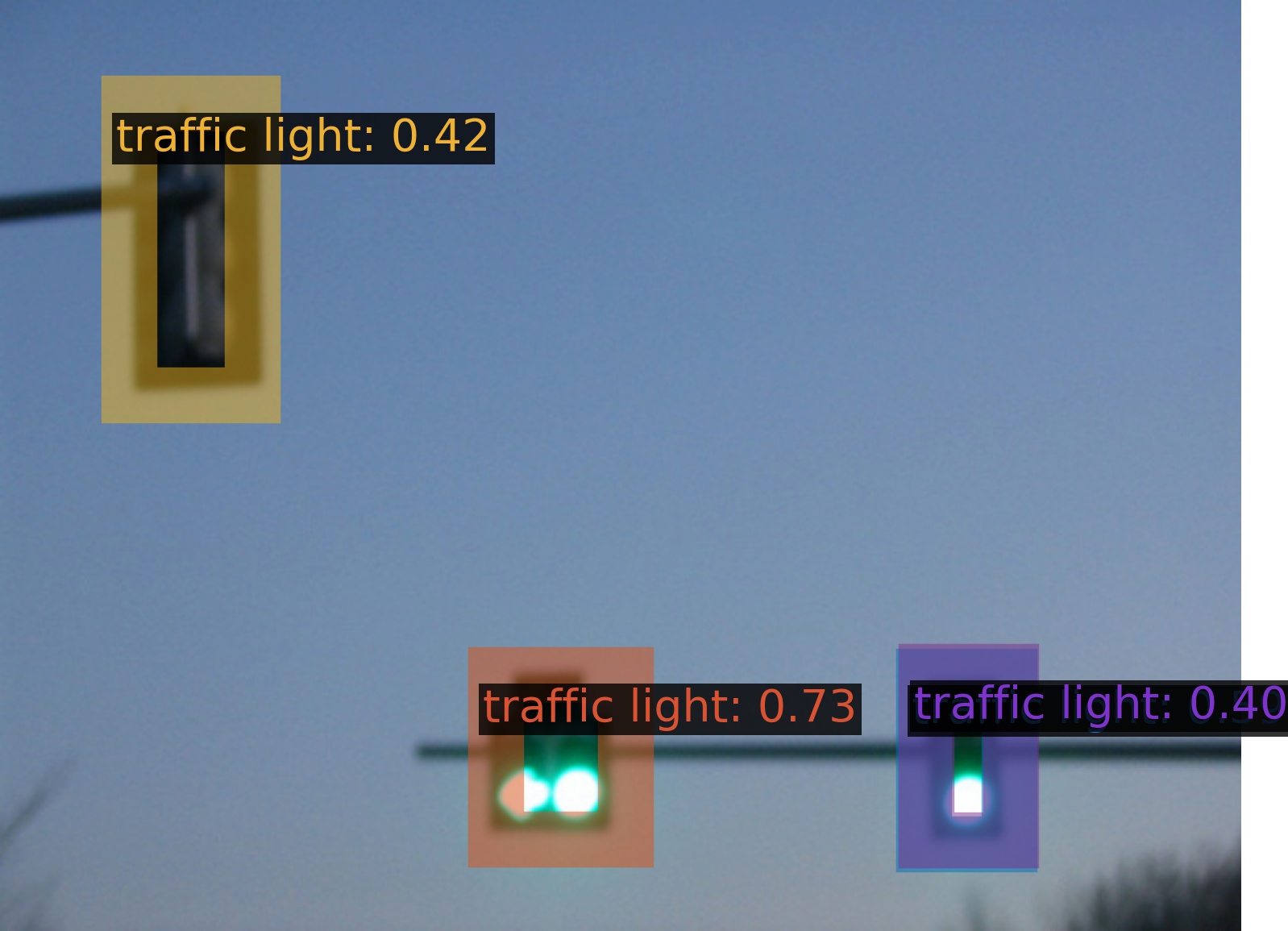} \\

    \end{tabular}}
\vspace{-0.25cm}
\caption{\textbf{Illustration showing the evolution of predictions of the proposed OW-DETR in the OWOD setting on MS-COCO images.} On the left: The objects detected by our OW-DETR when trained only on Task 1 classes. On the right: predictions for same images after incrementally training with Task 2 classes. In the top row, although the unknown objects (\textit{giraffe} and \textit{zebra}) are localized accurately, they are confused as a known class (\textit{horse}) during Task 1. However, these are corrected to their actual labels when trained incrementally in Task 2. In the bottom row, despite being localized not so accurately, multiple \textit{traffic lights} are correctly predicted as unknown class in Task 1 and these are detected accurately in Task 2. These results show promising performance of our OW-DETR in initially detecting potential unknown objects and later correctly detecting them when their corresponding classes are incrementally introduced.}
\label{fig:incr_2}
\end{figure*}

\noindent\textbf{OWOD comparison:} Figs.~\ref{fig:qual_t1} and~\ref{fig:qual_t2} show qualitative comparisons between ORE~\cite{joseph2021towards} and our proposed OW-DETR on example images in MS-COCO test-set. For each example image, detections of ORE are shown on the left, while the predictions of our OW-DETR are shown on the right. In general, we observe that the proposed OW-DETR obtains improved detections for the unknown objects, in comparison to ORE. \Eg, in top row of Fig.~\ref{fig:qual_t1}, while ORE fails to detect the \textit{refrigerator} (unknown in Task 1) in the left part of the image as unknown, our OW-DETR correctly predicts it as unknown. Similarly, in Fig.~\ref{fig:qual_t2} (top row), ORE wrongly predicts \textit{traffic light} on a road sign that is a true unknown, whereas our OW-DETR correctly detects it as an unknown object. These results show that the proposed contributions (attention-driven pseudo-labeling, novelty classification and objectness branch) in OW-DETR enable better reasoning \wrt the characteristics of unknown objects leading towards a more accurate detection in the OWOD setting. \\
\noindent\textbf{Evolution of predictions:} Figs.~\ref{fig:incr_1} and~\ref{fig:incr_2} illustrate an evolution of predictions when evaluating the proposed OW-DETR in different tasks of the OWOD setting on MS-COCO images. For each image, the objects detected by our OW-DETR when trained only on Task-1 classes is shown on the left. Similarly, the predictions after incrementally training with Task 2 classes is shown on the right. In the top row of Fig.~\ref{fig:incr_1}, a \textit{parking meter} (unknown in Task 1) is correctly detected as an unknown object during Task 1 evaluation and is rightly predicted as known class (\textit{parking meter}) during Task 2 after learning it incrementally. In Fig.~\ref{fig:incr_2} (top row), the unknown objects (\textit{giraffe} and \textit{zebra}) are localized accurately but they are confused as a known class (\textit{horse}) during Task 1, which can be attributed to visual similarity of these unknown objects with the known class \textit{horse}. However, these are correctly classified when the OW-DETR is trained incrementally in Task 2 with \textit{giraffe}  and \textit{zebra} included as new known classes. In the bottom row of Fig.~\ref{fig:incr_2}, despite the localization not being accurately performed, multiple \textit{traffic lights} are correctly predicted as an unknown class in Task 1 and these are detected accurately in Task 2 after incremental learning. These results show promising performance of our OW-DETR in initially detecting likely unknown objects and later accurately detecting them when their respective classes are incrementally introduced during the continual learning process.

In summary, the quantitative and qualitative results together show the benefits of our proposed contributions in detecting unknown objects in an open-world setting, leading towards a more accurate OWOD detection.

\section{Societal Impact and Limitations\label{sec:soci_impact}}
The open-world learning is a promising real-world setting which incrementally discovers novel objects. However, situations can arise where a particular object or fine-grained category must not be detected due to privacy or legal concerns. Similarly, an incremental model should be able to \emph{unlearn} (or forget) certain attributes or identities (object types in our case) whenever required. Specific solutions to these problems are highly relevant and significant, however, beyond the scope of our current work. 

Although our results in Tab.~\ref{tab:sota_owod} demonstrate significant improvements over ORE in terms of Recall and mAP, the performances are still on the lower side due to the challenging nature of the open-world detection problem. We hope that this work will inspire further efforts on this challenging but practical setting.

\section{Additional Implementation Details}
The multi-scale feature maps extracted from the backbone are projected to feature maps with $256$-channels ($D$) using convolution filters and used as multi-scale input to deformable transformer encoder, as in~\cite{zhu2020deformable}. We use the PyTorch~\cite{paszke2019pytorch} library and eight NVIDIA Tesla V100 GPUs to train our OW-DETR framework.
In each task, the OW-DETR framework is trained for $50$ epochs and finetuned for 20 epochs during the incremental learning step. Following~\cite{zhu2020deformable}, we train our OW-DETR using the Adam optimizer with a base learning rate of $2 \times 10^{-4}$, $\beta_1 = 0.9$, $\beta_2 = 0.999$, and weight decay of $10^{-4}$. For finetuning during incremental step, the learning rate is reduced by a factor of $10$ and trained using a set of $50$ stored exemplars per known class.

\vfill\null
{\small
\bibliographystyle{ieee_fullname}
\bibliography{egbib}
}


\end{document}